\documentclass{article} 
\usepackage[preprint]{colm2026_conference}

\usepackage{microtype}
\usepackage{hyperref}
\usepackage{url}
\usepackage{booktabs}
\usepackage[table]{xcolor}

\usepackage{enumitem}
\usepackage{graphicx}
\usepackage{float}
\usepackage{amsmath}
\usepackage{amssymb}
\usepackage{cleveref}
\usepackage{xspace}
\usepackage{multirow}
\usepackage{wrapfig}
\usepackage{array}
\usepackage{xcolor}
\usepackage{pifont}
\usepackage{makecell}
\usepackage{subcaption}
\usepackage{algorithm}
\usepackage{algpseudocode}
\usepackage{tcolorbox}
\tcbuselibrary{breakable}
\usepackage{listings}
\usepackage{longtable}
\usepackage{booktabs}
\usepackage{amssymb}

\usepackage{lineno}

\definecolor{darkblue}{rgb}{0, 0, 0.5}
\hypersetup{colorlinks=true, citecolor=darkblue, linkcolor=darkblue, urlcolor=darkblue}
\definecolor{veronica-red}{RGB}{196,30,58}

\definecolor{myrowblue}{HTML}{EAF4FF}
\definecolor{myrowblue_1}{HTML}{D6EAF8}
\newcommand{\cmark}{\textcolor[HTML]{2E8B57}{\ding{51}}}
\newcommand{\xmark}{\textcolor[HTML]{C0392B}{\ding{55}}}

\title{OpenMobile: Building Open Mobile Agents with Task and Trajectory Synthesis}

\vspace{-0.10cm}
\author{\normalfont\textbf{Kanzhi Cheng}\textsuperscript{1\,2\,3}\,\textsuperscript{*}\qquad
\textbf{Zehao Li}\textsuperscript{4}\qquad
\textbf{Zheng Ma}\textsuperscript{2}\textsuperscript{$\dagger$}\qquad
\textbf{Nuo Chen}\textsuperscript{1}\qquad
\textbf{Jialin Cao}\textsuperscript{1} \\
\textbf{Qiushi Sun}\textsuperscript{5}\qquad
\textbf{Zichen Ding}\textsuperscript{4}\qquad
\textbf{Fangzhi Xu}\textsuperscript{3\,6} \qquad
\textbf{Hang Yan}\textsuperscript{6}\qquad
\textbf{Jiajun Chen}\textsuperscript{1}\\
\textbf{Luu Anh Tuan}\textsuperscript{3}\qquad
\textbf{Jianbing Zhang}\textsuperscript{1}\textsuperscript{$\ddagger$}\qquad
\textbf{Lewei Lu}\textsuperscript{2}\textsuperscript{$\ddagger$}\qquad
\textbf{Dahua Lin}\textsuperscript{2}\\[0.5em]
\textsuperscript{1}Nanjing University\qquad
\textsuperscript{2}SenseTime\qquad
\textsuperscript{3}Nanyang Technological University\\
\textsuperscript{4}Shanghai AI Laboratory
\textsuperscript{5}The University of Hong Kong
\textsuperscript{6}Xi'an Jiaotong University
}

\begin{document}

\ifcolmsubmission
\linenumbers
\fi

\maketitle
\begingroup
\renewcommand\thefootnote{}
\footnotetext{
\footnotesize
\textsuperscript{*}Work done during internship at SenseTime.
\textsuperscript{$\dagger$}Proj leadership.
\textsuperscript{$\ddagger$}Corresponding author.
}
\endgroup

\vspace{-0.50cm}
\begin{abstract}
\vspace{-0.20cm}
Mobile agents powered by vision-language models have demonstrated impressive capabilities in automating mobile tasks, with recent leading models achieving a marked performance leap, e.g., nearly 70\% success on AndroidWorld.
However, these systems keep their training data closed and remain opaque about their task and trajectory synthesis recipes.
We present \textbf{OpenMobile}, an open-source framework that synthesizes high-quality task instructions and agent trajectories, with two key components:
(1) The first is a scalable task synthesis pipeline that constructs a global environment memory from exploration, then leverages it to generate diverse and grounded instructions.
and (2) a policy-switching strategy for trajectory rollout. By alternating between learner and expert models, it captures essential error-recovery data often missing in standard imitation learning.
Agents trained on our data achieve competitive results across three dynamic mobile agent benchmarks: notably, our fine-tuned Qwen2.5-VL and Qwen3-VL reach 51.7\% and \textbf{64.7\% on AndroidWorld}, far surpassing existing open-data approaches.
Furthermore, we conduct transparent analyses on the overlap between our synthetic instructions and benchmark test sets, and verify that performance gains stem from broad functionality coverage rather than benchmark overfitting.
We release data and code at \href{https://njucckevin.github.io/openmobile/}{OpenMobile} to bridge the data gap and facilitate broader mobile agent research.

\end{abstract}

\begin{figure}[h]
  \centering
  \includegraphics[width=\textwidth]{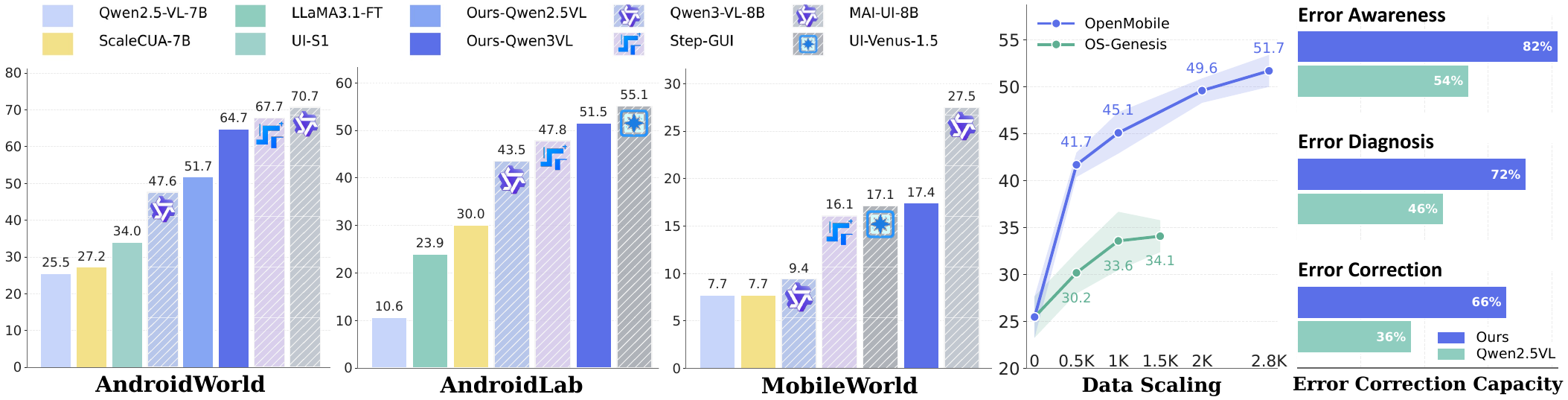}
  \caption{\textbf{Performance Comparison.} Task success rates across three dynamic mobile agent benchmarks. Our models significantly surpass open-data baselines and are competitive with leading closed-data systems. \textbf{Data Scaling.} AndroidWorld performance with increasing synthesized instructions. \textbf{Error Correction Capacity.} OpenMobile data substantially enhances the agent's error-recovery ability in live environments.}
  \label{fig:main}
\end{figure}

\vspace{-0.20cm}
\section{Introduction}
Vision-language models (VLMs) have fueled rapid progress in mobile agents—autonomous systems that interact with smartphone interfaces to complete user tasks.
Recent industry efforts such as Step-GUI \citep{yan2025step}, MAI-UI \citep{zhou2025mai}, UI-Venus-1.5 \citep{gao2026ui}, and MobileAgent-v3.5 \citep{xu2026mobile} have pushed the state of the art to striking levels, e.g., approaching 70\% task success on the widely-adopted AndroidWorld benchmark \citep{rawles2024androidworld}.
These advances are innately driven by large-scale, high-quality agent training data, consisting of task instructions paired with execution trajectories.

However, these leading systems uniformly keep their trajectory data closed and remain opaque about how such data are produced.
Meanwhile, the open-source community relies on public datasets like AndroidControl~\citep{li2024effects} and AMEX~\citep{chai2025amex}, 
achieves only around 30\% on the same benchmark \citep{lu2025ui,liu2025scalecua}.
Beyond this widening performance gap, the opacity prevents the community from understanding what data properties drive strong performance and generalization—leaving the open-source community unable to study, reproduce, or build upon these advances.

To bridge this gap,
we present OpenMobile, an open data synthesis framework and dataset for mobile agent training.
OpenMobile addresses two core challenges: 
\textbf{(1) Generating diverse, high-quality task instructions at scale in dynamic mobile environments.}
Existing approaches typically \emph{couple} exploration with generation by using a single trajectory as context for an LLM to task curation~\citep{sun2025genesis, murty2024nnetnav}. This dependency limits diversity to what a single local trajectory reveals.
We instead \emph{decouple} the two stages: first exploring the environment to build a global environment memory of the app's functionalities, then drawing on short-term memory from neighboring screens and long-term memory from semantically related functionalities within the app to compose complex, multi-step instructions.
\textbf{(2) Collecting agent trajectories that yield effective training signal.}
Expert trajectory distillation enables learners to imitate ideal behavior but often fails to address recovery from errors, 
which leads to a notable performance gap at test time. 
While self-evolution can mitigate this mismatch, it often suffers from slow convergence and is bounded by the learner's current performance caps.
To this end, 
we further introduce a policy-switching strategy that alternates between the learner and expert models during rollout. We find that error-intervention switching, where a monitor detects deviations to trigger expert corrections,
contributes to synthesizing error-recovery demonstrations while maintaining successful task completion.

Using OpenMobile, we synthesize 2.8K task instructions with corresponding 34K action steps across 20 Android apps.
We conduct a comprehensive evaluation on three established online benchmarks, i.e., AndroidWorld, AndroidLab~\citep{xu2025androidlab}, and MobileWorld~\citep{kong2025mobileworld}.
Notably, our fine-tuned Qwen2.5-VL-7B and Qwen3-VL-8B achieve 51.7\% and 64.7\% task success on AndroidWorld and improve on the challenging MobileWorld from 9.4\% to 17.4\%, competitive with leading closed-data solutions and larger foundation models \citep{bai2025qwen3}.
Beyond raw performance, we address the community's growing concerns about potential data contamination through transparent experiments, confirming that our performance gains stem from broad functionality coverage and enhanced error-recovery capabilities rather than benchmark overfitting.
These findings offer the open-source community a concrete foundation for building competitive mobile agents.

Our contributions are summarized as follows:

\begin{itemize}[leftmargin=1.5em]
  \item We propose and open-source OpenMobile, a task and trajectory synthesis framework for mobile agents. It introduces decoupled task synthesis that builds global environment memory for instruction generation, and policy-switching trajectory rollout that captures corrective signals absent from expert distillation.
  \item We conduct comprehensive evaluation on three challenging dynamic benchmarks. Agents trained on our data achieve competitive performance with closed-data systems.
  \item We provide systematic analyses to examine data contamination risks and demonstrate that our performance gains are driven by broad functionality coverage and enhanced error-recovery capabilities, rather than benchmark overfitting.
\end{itemize}

\section{Related Work}

Building autonomous agents for digital automation is a long-standing goal in the AI and NLP community~\citep{branavan2009reinforcement, shi2017world, shaw2023pixels}. 
Recent breakthroughs in LLMs have dramatically accelerated progress in this direction, enabling agents to plan and act across mobile, web, and desktop environments and tackle increasingly complex tasks \citep{rawles2024androidworld, zhou2023webarena, xie2024osworld, sun2025scienceboard}.

\paragraph{Digital Agents with Vision-Language Models.}
Early work relied on LLMs to interact with structured interface representations such as accessibility trees~\citep{deng2023mind2web, gur2023real}, or built agentic frameworks that operate computers via coding~\citep{wu2024copilot, sun2024survey}.
The rapid progress of vision-language models has driven a shift toward end-to-end, vision-centric GUI agents~\citep{cheng2024seeclick, gou2024navigating, wu2024atlas, wu2025gui}. 
These agents take raw screenshots as input and complete tasks through human-like actions such as clicking and typing.
Among these, proprietary systems such as Operator \citep{openai2025operator} and Anthropic's Computer-Use~\citep{anthropic2024computeruse} stand out, achieving impressive performance by leveraging frontier foundation models \citep{yang2026symphony}.
Meanwhile, UI-TARS~\citep{qin2025ui,wang2025ui} set a milestone for open-weight agents through GUI pretraining, trajectory fine-tuning, and online reinforcement learning.
More recently, industry efforts~\citep{yan2025step,zhou2025mai,gao2026ui,xu2026mobile} have further pushed mobile agent performance, achieving a 70\% task success rate on AndroidWorld.
At the core of these advances are large-scale synthesized task instructions and agent trajectories; however, both the data and the underlying synthesis recipes remain undisclosed.

On the other side, the open-source community has fallen increasingly behind.
Human-annotated datasets such as AndroidControl \citep{li2024effects} and AMEX \citep{chai2025amex} have provided a foundation \citep{rawles2023androidinthewild, lu2025guiodyssey, yang2026gui,sun2025sentinel}.
However, these datasets contain significant annotation noise and lack rich thinking patterns.
Models trained on them, such as ScaleCUA \citep{liu2025scalecua} and UI-S1 \citep{lu2025ui}, plateau at roughly 30\% on AndroidWorld.
There is a pressing need for scalable, open-source data synthesis recipes that can close this divide.

\paragraph{GUI Data Synthesis.}
Human annotation of GUI agent data is costly and labor-intensive, motivating growing interest in automated synthesis of task instructions and action trajectories.
Early efforts adopt a \emph{task-driven} paradigm, using strong language models to propose tasks from seed instructions and application descriptions \citep{he2024webvoyager, lai2024autowebglm}.
While straightforward, this approach lacks grounding in real-world context, often producing generic, underspecified, or infeasible instructions.
This has motivated \emph{interaction-driven} methods that first explore the target environment and then synthesize environment-grounded instructions from the observed context.
A representative work is OS-Genesis \citep{sun2025genesis}, which proposes reverse task synthesis, using random-walk trajectories to retrospectively infer meaningful task instructions.
NNetNav \citep{murty2024nnetnav} efficiently constructs complex web demonstrations through the synergy of an exploration strategy and a pruning labeler that filters low-quality trajectories.
Subsequent efforts further advance this paradigm with more structured exploration strategies \citep{yangself, gandhi2025go, jiang2026treecua} and more sophisticated instruction generation pipelines \citep{xie2025agentsynth, pahuja2025explorer, ramrakhya2025scaling}.
Across these methods, exploration and instruction generation remain tightly \emph{coupled}: each instruction is derived from a single exploration trajectory, which bounds diversity to local observations.

Once task instructions are available, the next step is to collect high-quality agent trajectories.
A prevalent approach is expert distillation, where a strong agent model rolls out trajectories and a verifier model filters them for quality \citep{pan2024autonomous, sun2025genesis, lin2025cuarewardbench}.
An alternative line explores self-evolution, where the agent iteratively executes tasks and retrains on its own successful trajectories to bootstrap performance \citep{he2025openwebvoyager, qin2025ui}.
Recent efforts further co-generate verifiable evaluation scripts alongside instructions to facilitate RL training \citep{xue2026evocua}.

\begin{figure}[ht]
  \centering
  \includegraphics[width=\textwidth]{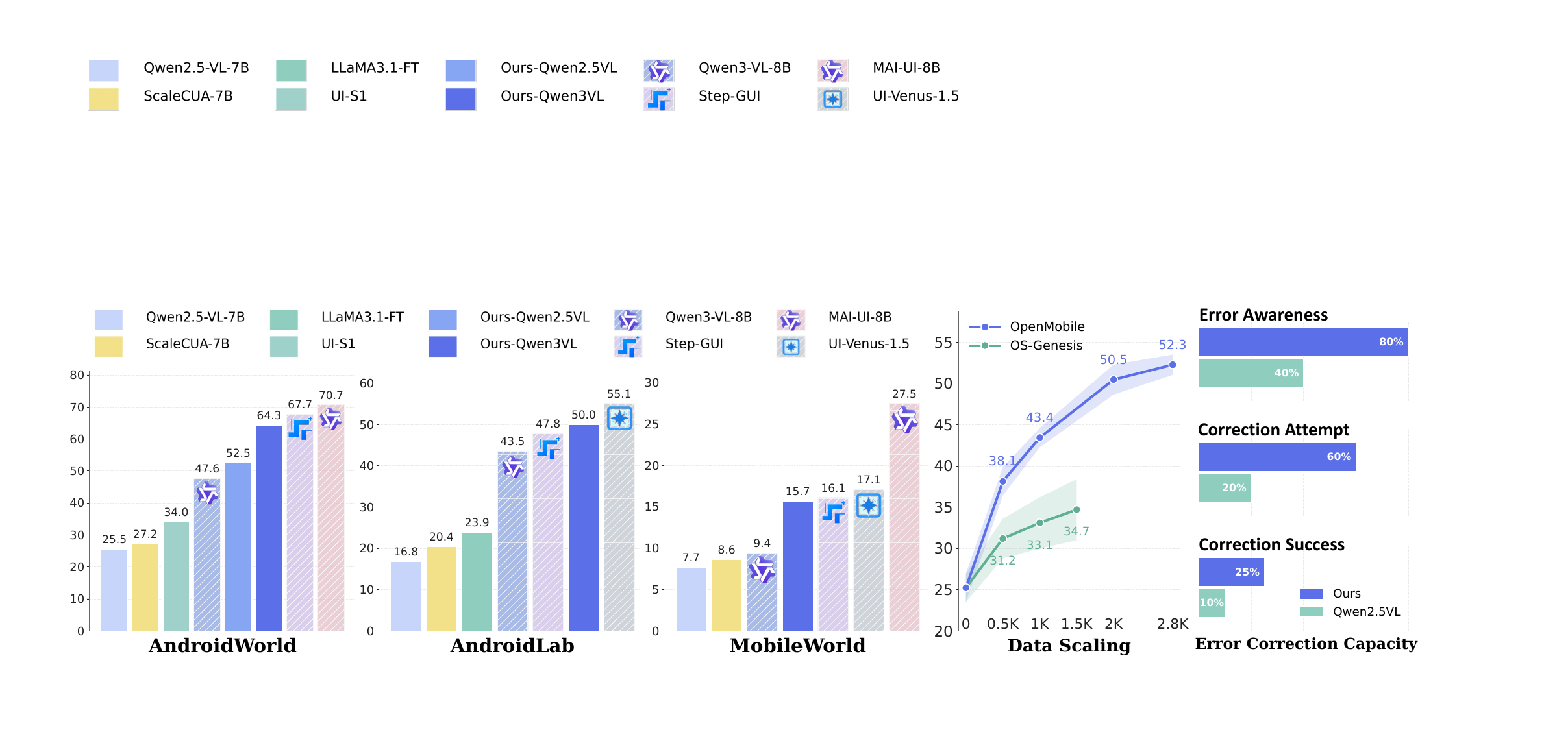}
  \caption{The overview of OpenMobile. (a) Scalable Task Synthesis. Instead of relying on a single local trajectory, we first explore the environment to build a global memory. By retrieving short- and long-term memories, we synthesize diverse, complex instructions that are contextually grounded and can be reasonably executed within the online environment. (b) Policy-Switching Rollout. Among different rollout strategies, error-intervention switching best captures error-recovery signals while ensuring task completion by detecting learner deviation and triggering expert correction.}
  \label{fig:fig2}
\end{figure}
\vspace{-0.10cm}

\section{OpenMobile}
The widening performance gap between proprietary and open-source mobile agents stems fundamentally from the lack of large-scale, high-quality open training data.
To bridge this divide, OpenMobile provides a data synthesis framework that produces two complementary assets: diverse, grounded task instructions over the broad functionalities of mobile environments (\Cref{sec:task_synthesis}), and agent execution trajectories enriched with error-recovery signals that facilitate efficient agent training (\Cref{sec:traj_rollout}).
The overview of our methods is in \Cref{fig:fig2}.
Implementation details are provided in \Cref{app:openmobile_detail}.

\vspace{-0.10cm}
\subsection{Scalable Task Synthesis}
\label{sec:task_synthesis}

Existing interaction-driven approaches tightly couple exploration with generation, deriving task instructions from individual exploration trajectories, which bounds instruction diversity to what one local trajectory reveals.
We instead propose a decoupled paradigm inspired by how humans learn a new application: one first explores to build a structured, comprehensive understanding of the app's capabilities, then recalls and composes relevant functionalities when a complex need arises.
Our pipeline mirrors this process in three stages: (i) exploring the environment to collect interaction experience, (ii) organizing the exploration data into a global environment memory $\mathcal{M}$, and (iii) drawing on short-term and long-term memory from $\mathcal{M}$ to synthesize compositional task instructions.

\paragraph{Environment Exploration}

The first stage aims to traverse target applications to collect information about the environment.
Through successive interactions with the app, this stage yields a set of exploration trajectories—sequences of screen-action interactions that capture the transitions between different states.
Our framework is agnostic to the specific exploration strategy and accommodates random walk~\citep{sun2025genesis}, structured coverage-based methods \citep{gandhi2025go, ramrakhya2025scaling, shao2026hats}, or human demonstrations.
To demonstrate that our approach does not rely on sophisticated exploration heuristics, we adopt a simple random walk in this work.
This choice is motivated by our observation that the critical factor lies not in exploration efficiency, but in how effectively the collected data is organized and used downstream.


\paragraph{Global Environment Memory Construction}

Multiple exploration sessions inevitably visit the same screen or environment state across different trajectories.
We exploit these shared screens as natural anchoring points to weave all fragmented trajectories into a unified, interconnected structure.
Specifically, we apply perceptual hashing to cluster visually similar screens, identifying a set of $N$ unique screens $\mathcal{S} = \{s_1, s_2, \ldots, s_N\}$, and aggregate transitions from all trajectories to form a neighborhood relation: for each screen $s_i$, its neighbors $\mathcal{N}(s_i) \subset \mathcal{S}$ are the screens directly reachable from or leading to $s_i$.
We then enrich each screen with its functionality set $\mathcal{F}(s_i) = \{f_1, f_2, \ldots, f_K\}$, extracted by a strong vision-language model, where each $f_k$ is a natural-language description capturing the semantics of a UI element on the screen.
To support cross-screen association, we compute semantic embeddings for all functionalities and build a per-app retrieval index.
The resulting global environment memory
\[
\mathcal{M} = \big(\,\mathcal{S},\;\mathcal{N},\;\{\mathcal{F}(s_i)\}_{i=1}^{N}\,\big)
\]
captures each application's functionality landscape in a structured, queryable form.

\paragraph{Memory-Augmented Task Synthesis}

Given the global memory $\mathcal{M}$, task synthesis proceeds by associating and composing functionalities into coherent instructions.
For each candidate screen $s_i$, we construct a context $\mathcal{C}(s_i)$ comprising three complementary views:
(1) the recalled screen $s_i$ itself, including its screenshot and annotated functionalities $\mathcal{F}(s_i)$, which serves as the focal point for generation;
(2) short-term memory $\mathcal{M_S}(s_i)$, the functionalities of neighboring screens $\mathcal{N}(s_i)$, reflecting what is locally reachable and naturally chainable with the current screen's capabilities;
and (3) long-term memory $\mathcal{M_L}(s_i)$, semantically related functionalities retrieved from distant screens within the same application, surfacing features that a user might associate through experience but that no single trajectory would reveal, encouraging cross-feature composition.

The full context $\mathcal{C}(s_i) = \big(s_i,\;\mathcal{M_S}(s_i),\;\mathcal{M_L}(s_i)\big)$, comprising both screenshots and textual descriptions, is presented to a VLM to generate task instructions based on this context.
We also carefully design generation guidelines and in-context examples to steer the model toward producing high-quality instructions; the full prompt is provided in the \Cref{app:prompt}.
The generated instructions undergo model-based quality filtering and embedding-based deduplication to yield the final instruction set.



\vspace{-0.10cm}
\subsection{Policy-Switching Rollout}
\label{sec:traj_rollout}

With task instructions in hand, the next step is to collect agent trajectories for training. A straightforward approach is expert distillation: rolling out a strong model to produce demonstration trajectories for imitation learning.
Although this yields high-quality trajectories, it restricts the learner to ideal behavior and fails to expose it to the mistakes it might make during inference, leaving the agent unable to recover from its own errors.
An alternative is self-evolution, where the learner iteratively executes tasks and retrains on its own successful trajectories. 
This directly addresses the distribution mismatch but converges slowly, as the learner's improvement is bounded by its own capacity.

We propose policy-switching rollout, which combines the strengths of both paradigms.
Given a task instruction $I$, an expert policy $\pi_e$, and a learner policy $\pi_l$, the rollout proceeds step by step: at each time step $t$, the agent observes the current screen $o_t$ and selects an action $a_t$ according to one of the two policies, controlled by a switching variable $z_t \in \{e, l\}$:
\[
a_t \sim \pi_{z_t}(\cdot \mid I, o_t, h_t)
\]
where $h_t$ denotes the interaction history up to step $t$.
By alternating between $\pi_e$ and $\pi_l$, 
the resulting trajectories contain segments where the learner makes mistakes followed by expert corrections.
This produces error-recovery experiences that are absent from pure distillation, while the expert's presence avoids the capacity ceiling inherent in self-evolution.

\vspace{-0.18cm}
\paragraph{Switching Strategies.}
The design of the switching rule, i.e., how $z_t$ is determined at each step, is critical.
A natural choice is random switching, which lets the learner take over with a fixed probability $p$ whenever the two policies disagree on the next action.
However, mobile agent tasks are inherently multi-solution: multiple valid action sequences can lead to the same goal, so disagreement between $\pi_e$ and $\pi_l$ does not necessarily signal a learner mistake, making random switching a noisy proxy for identifying errors.
Moreover, frequent switching between policies tends to disrupt coherent progress on complex tasks, resulting in fragmented trajectories that offer little usable training signal.

To address this, we design an error-intervention strategy.
Instead of switching at every point of disagreement, we introduce a monitor $\mathcal{O}$ that tracks the learner's execution in real time.
The rollout begins with the learner policy ($z_t = l$); only when $\mathcal{O}$ detects that the learner has deviated from productive progress does it trigger a switch to the expert ($z_t = e$) to intervene and correct the trajectory back on track.
The resulting trajectories are thus enriched with scarce error-recovery experiences, while expert intervention ensures sufficient task completion for effective training.
We compare these switching strategies in \Cref{sec:ablation}. 

\subsection{The OpenMobile Dataset}
\label{sec:dataset}

We instantiate the above pipeline on the Android emulator provided by AndroidWorld \citep{rawles2024androidworld}.
Although we leverage its environment infrastructure, we do \emph{not} incorporate any benchmark test instructions during synthesis to prevent data leakage; a detailed overlap analysis is provided in \Cref{sec:overlap}.
For policy-switching rollout, an early-stage fine-tuned checkpoint serves as the learner $\pi_l$ and \texttt{Gemini-3.1-Pro-Preview} as the expert $\pi_e$.
We adopt the action space and response format of Qwen3-VL~\citep{bai2025qwen3}, and employ the expert model to rewrite each step's chain-of-thought reasoning for higher-quality supervision.
The resulting dataset contains approximately 2,800 instructions and 34K corresponding action steps across 20 Android apps, with an average trajectory length of 12.2 steps and 129-word chain-of-thought reasoning per step.

\section{Experiments}

In this section, we train models on OpenMobile data and present main results on established dynamic benchmarks.
We leave ablation studies and further analyses to the next section.

\subsection{Experimental Settings}

From the policy-switching rollout trajectories, we retain only expert steps for training while preserving the full interaction history, including learner errors, as context to expose the model to realistic error-recovery scenarios.
We fine-tune two base models: Qwen2.5-VL-7B and Qwen3-VL-8B \citep{bai2025qwen3}.
The former has not undergone heavy GUI-specific post-training, offering a cleaner testbed for isolating data-driven gains; the latter serves as a stronger foundation to validate effectiveness on a more capable base and push the performance ceiling.
All models are trained with LLaMA-Factory \citep{zheng2024llamafactory} using standard supervised fine-tuning with a batch size of 32, a learning rate of 1e-5, for 3 epochs.

We also experiment with reinforcement learning (RL), including step-level RL \citep{lu2026ui} and trajectory-level agentic RL \citep{li2026themis}.
While the results prove that our synthesized trajectories remain effective, they do not yield significant improvements over SFT on dynamic benchmarks.
We discuss these findings in \Cref{app:rl}.

\subsection{Evaluation Benchmarks}

We evaluate on three established dynamic mobile agent benchmarks.
We focus on dynamic benchmarks because static datasets such as AndroidControl fundamentally lack the ability to evaluate an agent's crucial error-recovery capability, and their annotation noise further undermines evaluation reliability, making them poor proxies for real-world agent performance \citep{lu2025ui, gao2026ui}.
Experimental details are provided in \Cref{app:benchmark}.

\textbf{AndroidWorld} \citep{rawles2024androidworld} is the dominant mobile agent evaluation benchmark.
It provides robust, reproducible environments and deterministic evaluation through Android emulators. It comprises 116 tasks across 20 real-world apps, with parameterized task templates that generate diverse variants via random seeds.

\textbf{AndroidLab} \citep{xu2025androidlab} is a systematic benchmark for mobile agents with reproducible evaluation. 
It covers 138 tasks across 9 apps on predefined Android virtual devices, and supports both language-only and multimodal agents.

\textbf{MobileWorld} \citep{kong2025mobileworld} is a recent and more challenging dynamic benchmark.
It contains 201 tasks across 20 apps, with a stronger focus on long-horizon and cross-app workflows, making it substantially harder than AndroidWorld for evaluating complex mobile agent capabilities.
We evaluate on its GUI-only subset.

\renewcommand{\arraystretch}{1.1}
\begin{table*}[t]
\centering
\footnotesize
\resizebox{\linewidth}{!}{
\begin{tabular}{l|c|cc|cc|cc}
\toprule
\multirow{2}{*}{\textbf{Method}} & \multirow{2}{*}{\textbf{Base Model}} &
\multicolumn{2}{c|}{\textbf{AndroidWorld}} &
\multicolumn{2}{c|}{\textbf{AndroidLab}} &
\multicolumn{2}{c}{\textbf{MobileWorld}} \\
& 
& \textbf{Pass@1$\uparrow$} & \textbf{Pass@3$\uparrow$}
& \textbf{Pass@1$\uparrow$} & \textbf{Pass@3$\uparrow$}
& \textbf{Pass@1$\uparrow$} & \textbf{Pass@3$\uparrow$} \\
\midrule

\multicolumn{8}{c}{\cellcolor{gray!15} Commercial Models} \\
\midrule
GPT-4o & -- & 30.6 & -- & 31.2 & -- & -- & -- \\
Gemini-3-Pro & -- & 60.3 & 75.0 & -- & -- & 51.3 & -- \\
\midrule

\multicolumn{8}{c}{\cellcolor{gray!15} Open-Weight Models} \\
\midrule
\rowcolor{myrowblue}
Qwen2.5-VL-7B & -- & 25.5 ± 2.6 & 34.9 & 10.6 ± 1.8 & 15.2 & 7.7 ± 0.9 & 10.3 \\
\rowcolor{myrowblue_1}
Qwen3-VL-8B & -- & 47.6 ± 2.2 & 62.1 & 43.5 & -- & 9.4 & -- \\
UI-Venus-7B & {\fontsize{8}{9}\selectfont Qwen2.5-VL} & 49.1 & -- & 41.3 & -- & 8.5 & -- \\
Step-GUI-4B & {\fontsize{8}{9}\selectfont Qwen3-VL} & 63.9 & 75.8 & 47.8 & -- & 16.1 & -- \\
Step-GUI-8B & {\fontsize{8}{9}\selectfont Qwen3-VL} & 67.7 & 80.2 & -- & -- & -- & -- \\
MAI-UI-8B & {\fontsize{8}{9}\selectfont Qwen3-VL} & 70.7 & -- & -- & -- & 27.5 & -- \\
UI-Venus-1.5-8B & {\fontsize{8}{9}\selectfont Qwen3-VL} & 73.7 & -- & 55.1 & -- & 17.1 & -- \\
MobileAgent-v3.5-8B & {\fontsize{8}{9}\selectfont Qwen3-VL} & 71.6 & -- & --  & -- & 33.3 & --
\\
\midrule

\multicolumn{8}{c}{\cellcolor{gray!15} Open-Data Models} \\
\midrule
UI-S1-7B & {\fontsize{8}{9}\selectfont Qwen2.5-VL} & 34.0 & -- & -- & -- & -- & -- \\
ScaleCUA-7B & {\fontsize{8}{9}\selectfont Qwen2.5-VL} & 27.2 ± 2.2 & 36.2 & 30.0 ± 1.1 & 37.7 & 7.7 ± 0.4 & 8.6 \\
\rowcolor{myrowblue}
Ours-7B & {\fontsize{8}{9}\selectfont Qwen2.5-VL} & 51.7 ± 1.7 & 68.1 & 22.7 ± 0.4 & 37.0 & 14.8 ± 1.3 & 21.4 \\
\rowcolor{myrowblue_1}
Ours-8B & {\fontsize{8}{9}\selectfont Qwen3-VL} & 64.7 ± 3.2 & 78.0 & 51.5 ± 0.7 & 62.3 & 17.7 ± 2.2 & 24.8 \\
\bottomrule
\end{tabular}
}
\caption{Main results on AndroidWorld, AndroidLab, and MobileWorld. We report Pass@1 and Pass@3, where higher values indicate better performance. OpenMobile outperforms open-data baselines by a large margin and rivals leading closed-data systems.}
\label{tab:main_results}
\end{table*}

\subsection{Main Results}

\paragraph{OpenMobile data substantially improves mobile agent performance with strong generalization.}
As shown in \Cref{tab:main_results}, models fine-tuned on OpenMobile data significantly outperform corresponding baselines across all three benchmarks.
The Qwen2.5-VL variant improves by over 25 absolute points on AndroidWorld, demonstrating the effectiveness of our synthesized trajectories in enhancing VLM agentic capabilities.
Notably, although OpenMobile data is collected within the AndroidWorld environment, the resulting models generalize well to unseen settings, including novel apps in AndroidLab and long-horizon cross-app tasks in MobileWorld, e.g., achieving over 50\% relative improvement on the latter.
Overall, our models substantially surpass existing open-data approaches and are competitive with leading industrial efforts.
These results highlight the potential of open data synthesis for building competitive mobile agents.

\paragraph{Base model capability remains critical.}
Despite training on the same data, the Qwen3-VL variant consistently outperforms Qwen2.5-VL variant by a clear margin, suggesting that inherent base model capability, e.g., GUI understanding and planning, play an indispensable role.
While high-quality trajectory data can narrow the gap, improving the underlying foundation model remains equally important for pushing the performance ceiling.
Additional experiments with larger models and comparisons with other methods are in \Cref{additional_results}.

\section{Analysis}

In this section, we first ablate the key design choices behind OpenMobile (\Cref{sec:ablation}), and then investigate what drives its effectiveness (\Cref{sec:drive}), including the examination of potential benchmark overfitting.

\subsection{Ablation Study}
\label{sec:ablation}

\paragraph{OpenMobile produces diverse and high-quality instructions.}
We compare our decoupled task synthesis with OS-Genesis \citep{sun2025genesis} and a coupled baseline.
The coupled baseline shares the same generation prompt as ours, but uses the screenshot sequence from a single exploration trajectory as context instead of the global environment memory.
We first sample 50 instructions from each method for pairwise human evaluation, where experienced annotators judge instruction quality in terms of complexity and soundness and select the better one or declare a tie. As shown in \Cref{tab:ablation_task_human}, instructions synthesized by OpenMobile are notably more challenging than those from both baselines while maintaining comparable soundness.
We further train models on the synthesized data and evaluate on AndroidWorld.
Results in \Cref{tab:ablation_task_sr} show that OpenMobile achieves the best performance under a fixed budget of 1.5K trajectories.

\begin{table}[t]
\centering
\footnotesize
\setlength{\tabcolsep}{4pt}
\renewcommand{\arraystretch}{1.05}
\begin{minipage}[t]{0.58\textwidth}
\centering
\begin{tabular}{lcc}
\toprule
\textbf{OpenMobile vs.} & \textbf{Complexity} & \textbf{Soundness} \\
\midrule
OS-Genesis & 0.68 / 0.22 / 0.10 & 0.44 / 0.48 / 0.08 \\
Coupled Pipeline & 0.26 / 0.62 / 0.12 & 0.06 / 0.90 / 0.04 \\
\bottomrule
\end{tabular}
\subcaption{Human evaluation (win / tie / loss).}
\label{tab:ablation_task_human}
\end{minipage}
\hfill
\begin{minipage}[t]{0.38\textwidth}
\centering
\begin{tabular}{lc}
\toprule
\textbf{Method} & \textbf{Pass@1$\uparrow$} \\
\midrule
OS-Genesis & 34.1 ± 1.7 \\
Coupled Pipeline & 45.3 ± 2.2 \\
OpenMobile & \textbf{48.3 ± 1.3} \\
\bottomrule
\end{tabular}
\subcaption{Task success rate.}
\label{tab:ablation_task_sr}
\end{minipage}
\vspace{-0.15cm}
\caption{Ablation on task synthesis strategies. (a) Human evaluation of instruction quality over 50 pairwise comparisons. (b) AndroidWorld success rate with 1.5K trajectories.}
\vspace{-0.15cm}
\label{tab:ablation_task_synthesis}
\end{table}

\paragraph{Policy-switching rollout enriches error-recovery signals and boosts test-time performance.}
We compare four trajectory rollout strategies: (i) expert distillation, which collects trajectories
using only the expert model; (ii) self-evolution, where the learner iteratively retrains on its own successful trajectories over 3 rounds; (iii) random switching, which alternates between the expert and learner randomly as described in \Cref{sec:traj_rollout}; and (iv) our error-intervention switching.
Detailed experimental settings are provided in \Cref{app:policy_switching_setting}.

\begin{wraptable}{r}{0.48\textwidth}
\vspace{-0.2cm}
\centering
\footnotesize
\setlength{\tabcolsep}{4pt}
\renewcommand{\arraystretch}{1.05}
\begin{tabular}{lcc}
\toprule
\textbf{Rollout Strategy} & \makecell{\textbf{Avg.} \\ \textbf{ER}} & \textbf{Pass@1$\uparrow$} \\
\midrule
Expert Distillation & 0.42 & 44.8 ± 1.7 \\
Self-Evolution & 0.10 & 33.8 ± 0.9 \\
Random Switch & 0.64 & 45.1 ± 0.9 \\
Error-Intervention Switch & \textbf{1.56} & \textbf{48.3 ± 1.3} \\
\bottomrule
\end{tabular}
\caption{Ablation on rollout strategies. Avg. ER is the average number of error-recovery instances per trajectory, manually counted over 50 randomly sampled trajectories.}
\label{tab:ablation_rollout}
\vspace{-0.2cm}
\end{wraptable}

As shown in \Cref{tab:ablation_rollout}, error-intervention switching achieves the best downstream performance by introducing richer error-recovery signals during trajectory rollout.
Furthermore, as shown in the right panel of \Cref{fig:main}, we compare the error-recovery behaviors of our trained model and the base model during live execution, including error awareness, diagnosis, and correction.
The results confirm that OpenMobile data substantially strengthens the agent's error-recovery capability, which in turn drives the downstream performance gains.

\vspace{-0.15cm}
\begin{figure}[b!]
  \centering
  \includegraphics[width=0.85\textwidth]{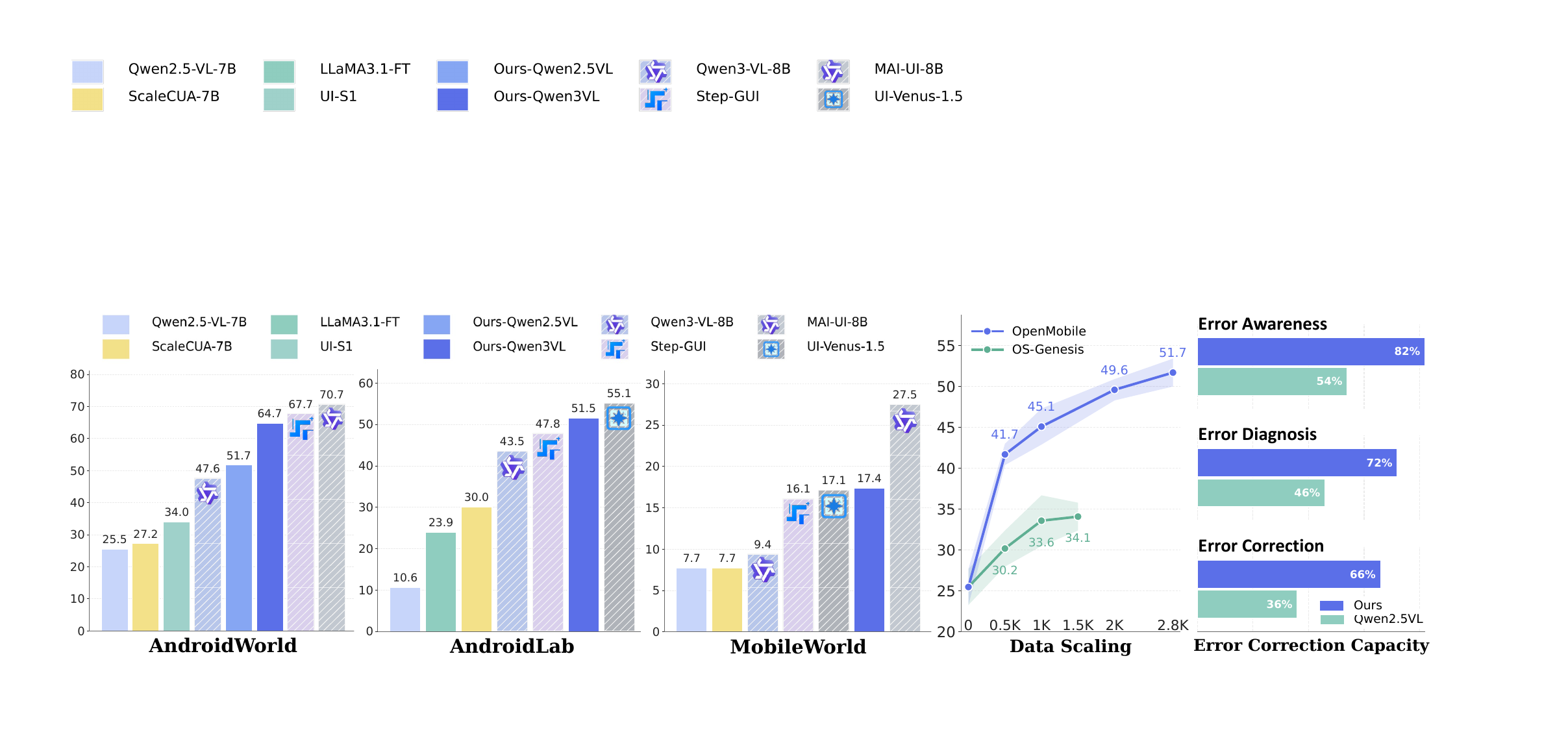}
  \caption{\textbf{Left:} Semantic similarity between synthetic and AndroidWorld instructions. Our synthesized instructions exhibit moderate functionality-level relevance, with only 3.5\% exceeding a similarity of 0.7. \textbf{Right:} Impact of removing test-similar instructions from training. Removing a small fraction of the most similar instructions causes only a marginal performance drop, mitigating benchmark overfitting concerns.}
  \label{fig:overlap}
\end{figure}

\subsection{What Drives the Effectiveness of OpenMobile Data?}
\label{sec:drive}

\paragraph{OpenMobile data is grounded in the benchmark environment, but does not overfit its test instructions.}
\label{sec:overlap}
Since our data is synthesized within the AndroidWorld environment, a natural concern is whether the instructions simply replicate benchmark tests.
To investigate, we compute semantic similarity between synthetic instructions and AndroidWorld test instructions using \texttt{openai/text-embedding-3-large}, and compare with AndroidControl and AMEX.
As shown in \Cref{fig:overlap} (left), OpenMobile instructions are indeed more similar to the test set, which is expected given the shared environment and app suite.
However, only 3.5\% of our instructions exceed a similarity of 0.7, indicating moderate relevance rather than task-level duplication and alleviating concerns about data leakage, e.g., through rephrasing test instructions.
The full list of most similar pairs is in \Cref{app:overlap_detail}.

Furthermore, we experiment with removing the most test-similar synthetic instructions and compare with random removal to observe the downstream performance impact.
As shown in \Cref{fig:overlap} (right), removing a small fraction, e.g., 10\%, leads to only a marginal drop, indicating that our gains do not fragily depend on a few test-similar samples.
However, as the removal ratio increases to 40\%, performance degrades notably compared to random removal.
This is because removing the most similar instructions inevitably strips away core app functionalities from the training data, preventing the model from acquiring essential skills.
We analyze the role of functionality coverage in detail in the following section.

\begin{figure}[t]
  \centering
  \includegraphics[width=0.85\textwidth]{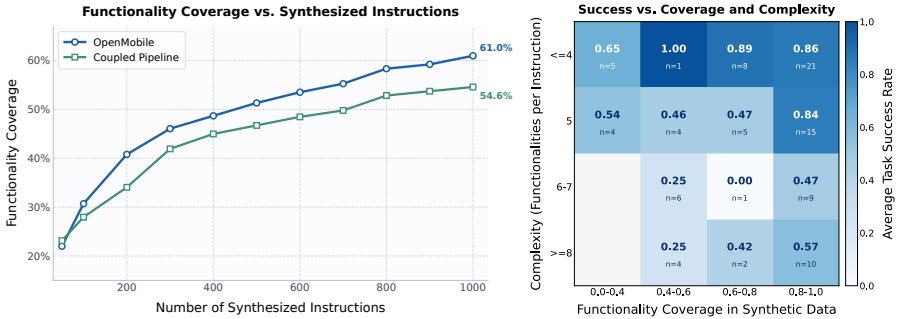}
  \caption{\textbf{Left:} Functionality coverage of AndroidWorld tasks as synthesized instructions scale. OpenMobile consistently achieves higher coverage than the coupled baseline. \textbf{Right:} Tasks with lower complexity (fewer required functionalities) and higher functionality coverage by synthetic data achieve higher success rates.}
  \label{fig:functionality}
\end{figure}

\paragraph{Broad functionality coverage drives agent performance.}
To understand what makes OpenMobile data effective, we conduct a functionality coverage analysis.
We use an LLM to decompose each test task into required atomic functionalities.
For instance, ``Create a calendar event titled Meeting with Team for tomorrow at 10am'' yields \textit{create calendar event}, \textit{set date}, \textit{set title}, and \textit{set start time}.
We then measure what fraction of test-required functionalities are covered by synthesized instructions.
As shown in \Cref{fig:functionality} (left), coverage steadily increases with instruction count, and OpenMobile consistently outperforms the coupled pipeline.
This confirms the advantage of our decoupled design: the global memory provides a structured view of the environment's capabilities, while retrieving semantically related functionalities as long-term context encourages cross-feature composition, together driving broader and more diverse instruction synthesis.

We further examine how task complexity, i.e., the number of atomic functionalities per task, and functionality coverage jointly affect success rate.
As shown in \Cref{fig:functionality} (right), tasks involving more functionalities are harder to complete (colors lighten from top to bottom), while tasks with higher coverage achieve higher success rates (colors deepen from left to right).
This highlights the importance of functionality coverage for instruction synthesis: an effective synthesis method should maximize coverage of the environment's core functionalities, which is precisely the design principle behind OpenMobile.

\section{Conclusion}

We presented OpenMobile, an open data synthesis framework for building competitive mobile agents.
Our framework addresses two key aspects of trajectory synthesis: (1) decoupling exploration from instruction generation to produce diverse, high-quality tasks, and (2) policy-switching rollout to enrich trajectories with error-recovery signals.
Agents trained on OpenMobile data achieve strong performance and generalize well to unseen dynamic environments, substantially narrowing the gap with closed-data industrial systems.
We also conduct transparent analyses on the overlap between synthetic and test instructions, confirming that these gains stem from broad functionality coverage and enhanced error-recovery capability rather than benchmark overfitting.
We release all data and code with the hope that OpenMobile serves as a foundation for broader open mobile agent research.



\bibliography{colm2026_conference}
\bibliographystyle{colm2026_conference}

\appendix
\crefalias{section}{appendix}

\section{Implementation Details of OpenMobile Framework}
\label{app:openmobile_detail}

\subsection{Environment Exploration Algorithm}

As described in \Cref{sec:task_synthesis}, OpenMobile decouples exploration from instruction generation, making the framework agnostic to the specific exploration strategy.
In this work, we follow OS-Genesis \citep{sun2025genesis} and adopt a simple random walk. Each session executes 10 steps, randomly selecting an interactable element from the current screen's accessibility tree to perform a click or type action.
A blacklist of non-interactable elements is maintained to avoid redundant interactions.
The primary objective of this phase is to maximize coverage of app states and functionalities for downstream task synthesis.

\subsection{Global Environment Memory Construction}
\label{app:memory_construction}

Given exploration trajectories consisting of screen-action transitions, we construct the global environment memory $\mathcal{M}$ through three stages: screen deduplication, functionality annotation, and semantic index construction. The full procedure is outlined in Algorithm~\ref{alg:memory_construction}.

\paragraph{Screen Deduplication.}
Exploration trajectories contain many visually identical or near-identical screens visited across different sessions. We compute a perceptual hash (pHash) for each screenshot and greedily cluster screens whose pHash similarity exceeds a threshold $\tau = 0.95$, selecting one representative per cluster. For each unique screen $s_i$, we aggregate all transitions from the original trajectories to identify its neighbor set $\mathcal{N}(s_i)$, i.e., screens directly reachable from or leading to $s_i$.

\paragraph{Functionality Annotation.}
For each unique screen, we use a strong LLM \texttt{Gemini-3.1-Pro-Preview} to extract a set of functionality descriptions $\mathcal{F}(s_i)$. Each functionality is a natural-language description capturing the semantics of a UI element (e.g., a button, toggle, or menu item). To improve annotation quality, we provide the model with the preceding screen and the action that led to the current screen as context. Elements are categorized as either \textit{functionality} (app-provided features such as buttons and toggles) or \textit{data} (user-generated content such as calendar events).

\paragraph{Semantic Index Construction.}
To enable cross-screen functionality retrieval, we compute semantic embeddings for all functionality descriptions within each app using a sentence embedding model \texttt{openai/text-embedding-3-large}. This produces a per-app retrieval index that supports efficient nearest-neighbor search during task synthesis. We apply greedy diversity filtering to ensure that retrieved functionalities are semantically distinct (pairwise cosine similarity below 0.8).

\begin{algorithm}[t]
\caption{Global Environment Memory Construction}
\label{alg:memory_construction}
\begin{algorithmic}[1]
\State \textbf{Input:}  Exploration trajectories $\mathcal{T} = \{(o_t, a_t, o_{t+1})\}$, similarity threshold $\tau$
\State \textbf{Output:} Global environment memory $\mathcal{M} = (\mathcal{S}, \mathcal{N}, \{\mathcal{F}(s_i)\})$

\Statex \textbf{// Stage 1: Screen Deduplication}
\State Collect all screens $\mathcal{O} = \{o_t, o_{t+1} \mid (o_t, a_t, o_{t+1}) \in \mathcal{T}\}$
\State Compute perceptual hash $h(o)$ for each $o \in \mathcal{O}$
\State $\mathcal{S} \gets \emptyset$
\For{each screen $o \in \mathcal{O}$}
    \If{$\nexists\, s \in \mathcal{S}$ s.t. $\textsc{Sim}(h(o), h(s)) \geq \tau$}
        \State $\mathcal{S} \gets \mathcal{S} \cup \{o\}$ \Comment{Add as new unique screen}
    \EndIf
\EndFor

\Statex \textbf{// Stage 2: Neighborhood \& Functionality Extraction}
\For{each unique screen $s_i \in \mathcal{S}$}
    \State $\mathcal{N}(s_i) \gets$ screens in $\mathcal{S}$ reachable from/to $s_i$ via transitions in $\mathcal{T}$
    \State $\mathcal{F}(s_i) \gets \textsc{VLM}(s_i, \text{context})$ \Comment{Extract functionality descriptions}
\EndFor

\Statex \textbf{// Stage 3: Semantic Index Construction}
\For{each app $A$}
    \State Collect all functionalities: $\mathcal{F}_A = \bigcup_{s_i \in A} \mathcal{F}(s_i)$
    \State Compute embeddings $\mathbf{E}_A = \textsc{Embed}(\mathcal{F}_A)$
    \State Build retrieval index over $(\mathcal{F}_A, \mathbf{E}_A)$
\EndFor

\State \Return $\mathcal{M} = (\mathcal{S}, \mathcal{N}, \{\mathcal{F}(s_i)\}_{i=1}^{|\mathcal{S}|})$
\end{algorithmic}
\end{algorithm}

\subsection{Memory-Augmented Task Synthesis}
\label{app:task_synthesis}

Given the global environment memory $\mathcal{M}$, we synthesize task instructions by presenting rich context to a strong vision-language model \texttt{Gemini-3.1-Pro-Preview} and prompting it to generate grounded, multi-step instructions.

\paragraph{Context Construction.}
For each candidate screen $s_i$, we construct a prompt that consists of three parts.
First, we include the screenshot of $s_i$ along with its annotated functionality descriptions, serving as the focal point for generation.
Second, we retrieve screenshots and functionality descriptions from neighboring screens in the transition graph as short-term memory. Specifically, we include 1 predecessor screen (the screen that transitions into $s_i$) and up to 3 successor screens (screens reachable from $s_i$), providing the model with local navigation context.
Third, we retrieve 30 semantically related functionalities from other screens within the same app as long-term memory. These are selected via embedding cosine similarity with a diversity constraint (pairwise similarity $< 0.8$) to surface distant but relevant features that encourage cross-functional composition.
The assembled context, comprising both screenshots and textual descriptions, is fed to the model together with generation guidelines and in-context examples.

\paragraph{Quality Filtering.}
Generated instructions undergo three-stage filtering:
(1) each instruction is scored by a strong LLM on complexity, clarity, and reasonableness (1--5 scale), and those with clarity $< 4$ or reasonableness $< 4$ are discarded;
(2) remaining instructions are sorted by scores and greedily deduplicated using embedding cosine similarity with a threshold of 0.8, so the highest-rated instruction is retained within each semantic cluster;
(3) per-app instruction counts are capped to ensure balanced app coverage during training.

\subsection{Error-Intervention Policy Switching}
\label{app:policy_switching_detail}

Our error-intervention switching begins with the learner model executing the task.
At each step, a monitor (\texttt{Gemini-3.1-Pro-Preview}) observes the recent action history and the last two screenshots to evaluate whether the previous action has caused the agent to deviate from the task objective.
Once deviation is detected, the expert model is invoked to intervene.
We find that providing the monitor's deviation analysis to the expert improves the quality of error-recovery signals, as the expert can better understand the current failure mode before correcting the trajectory.
In our rollout, this error-intervention process is triggered at most twice.
After each expert intervention, the expert executes at least 3 steps before control is returned to the learner.

\subsection{Prompts Used in OpenMobile}
\label{app:prompt}
We provide the complete prompts used in the OpenMobile pipeline below.

\begin{tcolorbox}[title=Prompt for Functionality Extraction, colback=gray!5, colframe=gray!50, fonttitle=\bfseries\small, fontupper=\small, fontlower=\small, breakable]

\textbf{System Prompt}
\vspace{0.3em}

You are a GUI screenshot analysis expert. You will be provided with:
\begin{enumerate}[leftmargin=1.5em, nosep]
    \item A screenshot of a UI screen (Screen Before) with the action area marked in red
    \item The action type performed
    \item The resulting screenshot after the action (Screen After)
    \item The name of the Android app
\end{enumerate}

Your task is to analyze the elements on the \textbf{second screenshot (Screen After)} ONLY. The first screenshot is provided only as context to help you understand the app's state.

Each element should be output as a dictionary:

\begin{verbatim}
{
    "type": "functionality" or "data",
    "label": "A short phrase describing its identifier on this screen",
    "description": "A few sentences describing this element's functionality"
}
\end{verbatim}

The description should be \textbf{comprehensive and detailed}:
\begin{itemize}[leftmargin=1.5em, nosep]
    \item Include the hierarchical location within the app (e.g., which menu, which settings page, which sub-section)
    \item Explain what this element does \textbf{at the phone/device level}, so that someone reading this description can fully understand the element's role and functionality without seeing the screenshot.
\end{itemize}

Here are examples showing bad descriptions and their improved versions:

\textit{Example 1:}\\
-- Bad: ``A WiFi toggle that enables or disables WiFi connectivity.''\\
-- Reason: Too vague; does not specify location or device-level changes.\\
-- Good: ``This toggle under System Settings > Network \& Internet > Wi-Fi enables or disables Wi-Fi on the device, allowing the phone to scan for available wireless networks and connect/disconnect from them.''

\textit{Example 2:}\\
-- Bad: ``A Reminder option enables users to set a reminder.''\\
-- Reason: Too vague; does not explain what scenario it is used for.\\
-- Good: ``In the calendar app's event creation/edit screen, this reminder option schedules a notification before the event starts (e.g., 10 minutes in advance), helping the user receive an alert at the chosen lead time.''

Output a JSON list only. No markdown, no comments, no extra text. Start with \texttt{[} and end with \texttt{]}.

\tcblower

\textbf{User Prompt}
\vspace{0.3em}

\texttt{[Image: Screen Before (with action area marked in red)]}\\
\texttt{[Image: Screen After]}

App: \{app\_name\}\\
Action: \{action\_type\}

The first image is Screen Before (with action area marked in red). The second image is Screen After. Please analyze the elements on the second image.

\end{tcolorbox}

\begin{tcolorbox}[title=Prompt for Task Instruction Synthesis, colback=gray!5, colframe=gray!50, fonttitle=\bfseries\small, fontupper=\small, fontlower=\small, breakable]

\textbf{System Prompt}
\vspace{0.3em}

You are a GUI explorer. Your goal is to explore a GUI environment and synthesize high-quality, high-difficulty, executable, high-level, multi-step GUI tasks/instructions.

You have already completed the exploration work. You have collected many screenshots from the current GUI environment, the transitions between them, and various functionalities within the corresponding app.

Now, you need to fully associate and imagine based on the following three sources of information to generate long-range, high-level tasks/instructions that are possible within the current app:
\begin{enumerate}[leftmargin=1.5em, nosep]
    \item A recalled screenshot of a specific screen
    \item Several screenshots in short-term memory that have transition relationships with this screenshot (screens that can be reached from the current screen)
    \item Importantly, some functionalities retrieved from long-term memory that are associated with the current screen (semantically related functionalities from other screens in the same app)
\end{enumerate}

Based on these three sources of information, you should fully associate, imagine, and generate long-range, high-level tasks/instructions that are possible within the current app.
\\

\textbf{Guidelines}

\begin{enumerate}[leftmargin=1.5em]
    \item The provided screenshots and functionalities are only a portion of your recalled memories serving as context. Your ONLY task is to synthesize clear multi-step GUI instructions. The instructions you synthesize do not need to have direct connections with the current screen or operations, but can be inferred from the context. However, to ensure the difficulty and complexity of generated tasks, you are encouraged to analyze, associate, and combine functionalities from your memories.

    \item There are two types of tasks to generate:
    \begin{itemize}[nosep]
        \item \textbf{Action tasks}: Require performing a series of actions to accomplish a goal. For example: ``Set an alarm for tomorrow at 8 AM that repeats every weekday.''
        \item \textbf{Question-answering tasks}: Require performing a series of actions and answering a question related to the environment's content. For example: ``In my to-do list, how many tasks need to be completed this Wednesday? Answer the question with a single number.''
    \end{itemize}
    You should decide which type of task is appropriate to generate based on the context.

    \item Synthesized tasks \textbf{must be clear and explicit}. Generated tasks should be specific with sufficient details, so that executors will not feel confused. For example, ``Help me create a new event in the calendar'' is too broad. It should include concrete configurations, e.g., date, time, title, description, duration, location, etc.

    \item Synthesized tasks must be executable. \textbf{If you want to generate a task that involves operating on app data (for example, deleting an entry in the calendar), you MUST make sure the data you want to operate on is present in the given screenshots.}

    \item Generated tasks should be diverse. Do not only focus on the app's main functions. Try to cover all functionalities of the app as much as possible, for example, elements or functions in corners of screens, or functionalities you associate from memories.

    \item Generated tasks should be long-range. Do not generate single-step tasks such as clicking a button. You are encouraged to generate tasks that require executors to reason, plan, and complete in multiple steps. \textbf{You can also consider combining different sub-functions or sub-tasks into a long-range task, but ensure reasonableness.}

    \item Generated tasks should be high-level. \textbf{Do not generate step-by-step instructions and detailed actions.} Instead, integrate multi-step instructions into a high-level intent to increase task difficulty. \textbf{They should be a single command that contains specific details, rather than step-by-step operations for completing a task.}

    \item Generated tasks should start from the phone's home screen, not from the currently provided screen. Do not generate tasks that are bound to temporary states of the current interface (for example, a popup dialog that appears).

    \item The operating environment is a virtual device with no network connection. Do not generate tasks that require internet connection or login. However, you can freely use data that is already saved in the existing app.
\end{enumerate}

\textbf{Example Tasks}

Here are examples showing bad tasks and their improved versions:

\textit{Example 1:}\\
-- Bad: ``Access and manage the list of all saved Bluetooth devices.''\\
-- Reason: Does not specify what ``manage'' means.\\
-- Good: ``View all existing Bluetooth devices, and if any exist, delete all of them.''

\textit{Example 2:}\\
-- Bad: ``Add a new recipe to the list using the plus button on the main recipe screen.''\\
-- Reason: Does not specify concrete content.\\
-- Good: ``In the Broccoli app, add a new recipe for `Tomato and Egg Stir-fry', set the category to `Stir-fry', and fill in the description as `Mom's favorite dish'.''

\textit{Example 3:}\\
-- Bad: ``Check the battery usage statistics and enable Battery Saver mode if necessary.''\\
-- Reason: ``If necessary'' will confuse the executor.\\
-- Good: ``Write the top three items from battery usage statistics into the Markor app and save it as `battery\_usage\_statistics', and enable Battery Saver mode.''

\textit{Example 4:}\\
-- Bad: ``Dismiss the voice search connection error by tapping the `Keyboard' button, then manually type `The Beatles' in the search bar.''\\
-- Reason: Includes a temporary state and assumes starting from the search interface.\\
-- Good: ``In \{app name\}, how many songs are included for The Beatles and Taylor Swift respectively? Answer with numbers separated by a comma.''

\textit{Example 5:}\\
-- Bad: ``In the Broccoli app, use the search function to find the recipe `Salmon with Dill Sauce'. Open its details page and answer how many servings it yields.''\\
-- Reason: Contains too many specific operations; should be more high-level.\\
-- Good: ``In the Broccoli app, how many servings does `Salmon with Dill Sauce' provide, and what is the total preparation time required?''

\textit{Example 6:}\\
-- Bad: ``In Simple Calendar Pro, navigate to the `Customize colors' menu, attempt to change the App icon color, and dismiss the warning popup.''\\
-- Reason: Contains unnecessary specific operations and temporary states.\\
-- Good: ``Set the app color of Simple Calendar Pro to blue.''

\textit{Example 7:}\\
-- Bad: ``In the Tasks app, what tasks do I have?''\\
-- Reason: Too vague.\\
-- Good: ``In the Tasks app, which tasks due this week are not completed yet? Answer with titles only; if there are multiple, separate them with commas.''

\textit{Example 8:}\\
-- Bad: ``In the Audio Recorder app, configure the settings for high-fidelity recording. After entering the app, navigate to the setup menu and change the recording format to Wav, set the sample rate to 48kHz...''\\
-- Reason: Contains too many step-by-step operations.\\
-- Good: ``Record an audio file in Wav format with 48kHz sample rate and Stereo channel using Audio Recorder, and save it as test\_audio.''

\tcblower

\textbf{User Prompt}
\vspace{0.3em}

\texttt{[Image: Current Screen]}\\
\texttt{[Image: Preceding Screen 1 (if available)]}\\
\texttt{[Image: Associated Screen 1--3 (if available)]}

\#\# Current Screen\\
\textbf{App}: \{app\_name\}\\
\textbf{Elements on Current Screen} (\{N\} items):\\
\hspace{1em}1. ``type'': functionality, ``description'': \{description\_1\}\\
\hspace{1em}2. ...

These are screens that can transition into the current screen:\\
\#\#\# Preceding Screen 1\\
\textbf{Elements} (\{N\} items): ...

These are screens that can be reached from the current screen:\\
\#\#\# Associated Screen 1\\
\textbf{Elements} (\{N\} items): ...

\#\# Related Functionalities from Other Screens (\{M\} items)\\
These are semantically related functionalities from other screens in the same app:\\
\hspace{1em}1. \{description\_1\}\\
\hspace{1em}2. ...

\#\# Your Task\\
Based on the above context, carefully analyze and think, then generate 1--3 high-quality GUI tasks. Each task should be a concise but high-level instruction in English. Output format (JSON array):
\begin{verbatim}
[
  {"reasoning": "...", "task": "task instruction 1"},
  {"reasoning": "...", "task": "task instruction 2"}
]
\end{verbatim}

\end{tcolorbox}

\section{Experiment Settings}

\subsection{Benchmark Evaluation Setup}
\label{app:benchmark}

We evaluate on three established dynamic mobile agent benchmarks: AndroidWorld, AndroidLab, and MobileWorld.
All models are deployed with vLLM for inference.
We observe that agent execution in dynamic environments exhibits inherent randomness, leading to non-trivial variance in success rates across runs.
To account for this, we run each benchmark three times and report the mean along with the half-range (i.e., (max $-$ min) / 2) as a measure of variation.
We additionally report Pass@3, the success rate when a task is considered solved if any of the three runs succeeds, to indicate the model's performance upper bound.
Results of some baseline models are taken from prior work, e.g., UI-Venus-1.5 \citep{gao2026ui} and the MobileWorld leaderboard \citep{kong2025mobileworld}.

\subsection{Policy-Switching Rollout Settings}
\label{app:policy_switching_setting}
We use Qwen2.5-VL-7B-Instruct as the base model (learner $\pi_l$) for all policy-switching ablations, with \texttt{Gemini-3.1-Pro-Preview} as the expert $\pi_e$.
\textbf{Expert distillation.} The expert model executes all synthesized instructions. We retain trajectories where the expert signals task completion (i.e., outputs \texttt{complete} or \texttt{answer}) and convert them into step-level training samples.
\textbf{Self-evolution.} The learner executes the synthesized instructions, and the expert serves as a judge to identify successful trajectories. Only successful trajectories are used to retrain the learner. This process is iterated for 3 rounds.
\textbf{Random switching.} At each step, whenever the learner and expert predict inconsistent actions (e.g., different action types or different target elements), the learner's action is used in place of the expert's. However, the learner is not allowed to execute terminal actions (\texttt{complete} or \texttt{answer}) to ensure task completion by the expert.
\textbf{Error-intervention switching.} The rollout begins with the learner policy. A monitor tracks the learner's execution and triggers a switch to the expert when deviation from productive progress is detected. The expert then intervenes to correct the trajectory back on track.
Details of the monitor design are provided in \Cref{app:policy_switching_detail}.
All strategies are compared under a fixed budget of 1.5K trajectories.
To quantitatively measure the error-recovery signals introduced by each strategy, we randomly sample 50 trajectories from each and manually inspect the average number of error-recovery signals per trajectory, defined as a step where the agent recognizes and attempts to correct a mistake from the previous step.

\section{Exploration with Reinforcement Learning}
\label{app:rl}
Beyond standard supervised fine-tuning (SFT), we explore the effectiveness of reinforcement learning (RL) on our synthesized data.

\paragraph{Step-Level RL.}
We first experiment with step-level RL, a commonly adopted paradigm for GUI agents.
Following UI-R1~\citep{lu2025ui}, we define three reward signals for each agent step: a format reward, an action type reward, and a grounding reward, and train with standard GRPO.
We then evaluate the resulting model on AndroidWorld.
Results show that while step-level GRPO initially improves performance using synthesized trajectories, the gains quickly saturate and ultimately fail to surpass the SFT baseline.
We attribute this to the inherent discrepancy between step-level optimization and multi-step execution in dynamic environments.
Step-level RL tends to overfit to single-step outputs, failing to produce stable gains on long-horizon tasks that require sustained interaction with a changing environment.
A similar observation has also been reported by UI-Venus-1.5~\citep{gao2026ui}.

\renewcommand{\arraystretch}{1.1}
\begin{table*}[b]
\centering
\footnotesize
\resizebox{\linewidth}{!}{
\begin{tabular}{l|c|cc|cc|cc}
\toprule
\multirow{2}{*}{\textbf{Method}} & \multirow{2}{*}{\textbf{Base Model}} &
\multicolumn{2}{c|}{\textbf{AndroidWorld}} &
\multicolumn{2}{c|}{\textbf{AndroidLab}} &
\multicolumn{2}{c}{\textbf{MobileWorld}} \\
& 
& \textbf{Pass@1$\uparrow$} & \textbf{Pass@3$\uparrow$}
& \textbf{Pass@1$\uparrow$} & \textbf{Pass@3$\uparrow$}
& \textbf{Pass@1$\uparrow$} & \textbf{Pass@3$\uparrow$} \\
\midrule

Qwen3-VL-8B & -- & 47.6 ± 2.2 & 62.1 & 43.5 & -- & 9.4 & -- \\
Ours-8B & {\fontsize{8}{9}\selectfont Qwen3-VL} & 64.7 ± 3.2 & 78.0 & 51.5 ± 0.7 & 62.3 & 17.7 ± 2.2 & 24.8 \\
Ours-8B-RL & {\fontsize{8}{9}\selectfont Qwen3-VL} & 64.1 ± 0.5 & 77.6 & 53.9 ± 1.5 & 63.0 & 16.8 ± 0.9 & 20.5 \\
\bottomrule
\end{tabular}
}
\caption{Results of trajectory-level RL.}
\label{tab:app_rl}
\end{table*}

\paragraph{Trajectory-Level Agentic RL.}
We further explore trajectory-level agentic RL to enhance the agent's performance.
We conduct these experiments using the OS-Themis~\citep{li2026themis} framework, which provides an infrastructure of over a hundred Android emulator instances for trajectory rollout, along with a multi-agent critic that leverages VLMs to judge task success or failure.
We use an early checkpoint as the starting point for RL training, and filter the synthesized task instructions to retain only those that the checkpoint fails but the expert successfully completes, yielding 244 instructions in total.
As shown in \Cref{tab:app_rl}, while trajectory-level RL on our synthesized instructions does improve performance, it cannot consistently surpass its fully SFT-trained counterpart.
We hypothesize that this is related to the limited diversity of our current environment setup and the stability of the RL framework itself.
Addressing these limitations to further advance mobile agent capabilities remains an important direction for future work.

\section{Additional Experimental Results}
\label{additional_results}

\begin{table}[ht]
\centering
\small
\setlength{\tabcolsep}{4pt}
\renewcommand{\arraystretch}{1.05}
\begin{minipage}[t]{0.48\textwidth}
\centering
\begin{tabular}{lcc}
\toprule
\textbf{Method} & \textbf{Pass@1$\uparrow$} & \textbf{Pass@3$\uparrow$} \\
\midrule
Qwen2.5-VL-7B & 25.5 ± 2.6 & 34.9 \\
Qwen2.5-VL-72B & 27.6 & -- \\
UI-Venus-72B & 65.9 & -- \\
\midrule
Ours-7B & 51.7 ± 1.7 & 68.1 \\
Ours-72B & 59.3 ± 0.9 & 72.8 \\
\bottomrule
\end{tabular}
\subcaption{Scaling to larger models.}
\label{tab:appendix_scaling}
\end{minipage}
\hfill
\begin{minipage}[t]{0.48\textwidth}
\centering
\begin{tabular}{lccc}
\toprule
\textbf{Method} & \textbf{Open} & \textbf{\#Traj} & \textbf{Pass@1$\uparrow$} \\
\midrule
OS-Genesis & \cmark & 1.5K & 17.4 \\
HATS & \cmark & 1K & 24.4 \\
AutoPlay & \xmark & 20K & 40.1 \\
MobileGen & \xmark & 0.5K & 45.7 \\
OpenMobile & \cmark & 2.8K & \textbf{64.7} \\
\bottomrule
\end{tabular}
\subcaption{Comparison with data synthesis methods.}
\label{tab:appendix_comparison}
\end{minipage}
\caption{(a) AndroidWorld results with larger model sizes. (b) Comparison with existing data synthesis methods on AndroidWorld.}
\label{tab:appendix_additional}
\end{table}

To validate the effectiveness of OpenMobile data on larger models, we fine-tune \texttt{Qwen2.5-VL-72B-Instruct} using the same data. As shown in \Cref{tab:appendix_scaling}, the larger model yields notably stronger performance, confirming both the scalability of our data and the importance of base model capability.

We also compare with existing mobile agent data synthesis methods \citep{sun2025genesis, shao2026hats, ramrakhya2025scaling, kang2026learning}.
As shown in \Cref{tab:appendix_comparison}, OpenMobile achieves a substantially higher success rate with a moderate data scale.
We note that direct comparison is imperfect, as these methods differ in base models and experimental settings, and some do not open-source full implementation details.
Nevertheless, the results demonstrate the effectiveness of OpenMobile and position it as a strong starting point for future mobile agent data synthesis research.

\section{Similarity Between Synthetic and Test Instructions}
\label{app:overlap_detail}

To quantify the overlap between our synthesized instructions and the AndroidWorld test set, we compute pairwise cosine similarities using sentence embeddings from \texttt{openai/text-embedding-3-large}.
As reported in \Cref{fig:overlap}, our synthesized instructions exhibit moderate functionality-level relevance rather than task-level overlap with the benchmark, with only 3.5\% exceeding a similarity of 0.7.
\Cref{tab:overlap_full} lists each AndroidWorld test instruction alongside its most similar synthetic counterpart from synthesized instructions.

\begin{scriptsize}
\setlength{\tabcolsep}{3pt}
\begin{longtable}{@{} c >{\raggedright\arraybackslash}p{0.44\textwidth} >{\raggedright\arraybackslash}p{0.44\textwidth} c @{}}
\caption{Complete list of nearest synthetic--test instruction pairs. Each row shows an AndroidWorld test instruction paired with the most similar OpenMobile synthetic instruction.}
\label{tab:overlap_full} \\
\toprule
\textbf{\#} & \textbf{AndroidWorld Test Instruction} & \textbf{Most Similar Synthetic Instruction} & \textbf{Sim.} \\
\midrule
\endfirsthead

\multicolumn{4}{c}{\scriptsize\itshape (continued from previous page)} \\[0.5ex]
\toprule
\textbf{\#} & \textbf{AndroidWorld Test Instruction} & \textbf{Most Similar Synthetic Instruction} & \textbf{Sim.} \\
\midrule
\endhead

\midrule
\multicolumn{4}{r}{\scriptsize\itshape (continued on next page)} \\
\endfoot

\bottomrule
\endlastfoot

1 & Record an audio clip using Audio Recorder app and save it. & In the Audio Recorder app, change the recording settings to use the Wav format and Stereo channel, then record a short audio clip. & 0.715 \\
\addlinespace[2pt]
2 & Record an audio clip and save it with name "eVq3\_review.m4a" using Audio Recorder app. & Record a high-quality audio clip using the Wav format and 48kHz sample rate in the Audio Recorder app. & 0.635 \\
\addlinespace[2pt]
3 & Open the file task.html in Downloads in the file manager; when prompted open it with Chrome. Then create a drawing using the three colors shown at the top and hit submit. & In the Files app, navigate to the Downloads folder and open the 'task.html' file using the Chrome browser. & 0.743 \\
\addlinespace[2pt]
4 & Open the file task.html in Downloads in the file manager; when prompted open it with Chrome. Then navigate the X to the bottom-right cell, by using the direction buttons. & In the Files app, navigate to the Downloads folder and open the 'task.html' file using the Chrome browser. & 0.803 \\
\addlinespace[2pt]
5 & Open the file task.html in Downloads in the file manager; when prompted open it with Chrome. Then click the button 5 times, remember the numbers displayed, and enter their product in the form. & In the Files app, navigate to the Downloads folder and open the 'task.html' file using the Chrome browser. & 0.706 \\
\addlinespace[2pt]
6 & Take one photo. & Switch to Camera mode, enable the 3x3 grid lines overlay, and take a photo. & 0.524 \\
\addlinespace[2pt]
7 & Take one video. & Switch the Camera app to Video mode and record a short video clip. & 0.502 \\
\addlinespace[2pt]
8 & Pause the stopwatch. & Use the Stopwatch to record a lap time, then pause and reset the timer to zero. & 0.626 \\
\addlinespace[2pt]
9 & Run the stopwatch. & Use the Stopwatch to record a lap time, then pause and reset the timer to zero. & 0.610 \\
\addlinespace[2pt]
10 & Create a timer with 23 hours, 4 minutes, and 57 seconds. Do not start the timer. & In the Clock app, set a timer for 1 hour, 23 minutes, and 45 seconds and start the countdown. & 0.574 \\
\addlinespace[2pt]
11 & Create a new contact for Ahmed dos Santos. Their number is +12432810546. & Create a new contact for 'Alice Smith' with the phone number '555-123-4567' in the Phone app. & 0.527 \\
\addlinespace[2pt]
12 & Go to the new contact screen and enter the following details: First Name: Eva, Last Name: Smith, Phone: 119-168-9838, Phone Label: Work. Do NOT hit save. & In the Contacts app, create a new contact with the name "John Smith" and the phone number "555-1234". & 0.614 \\
\addlinespace[2pt]
13 & Add the following expenses into the pro expense: name\textbar{}amount\_dollars\textbar{}category\_name\textbar{}note Museum Tickets\textbar{}\$325.17\textbar{}Entertainment\textbar{}Urgent Social Club Dues\textbar{}\$425.35\textbar{}Social\textbar{}I may repeat this Museum Tickets\textbar{}\$485.01\textbar{}Entertainment\textbar{}I may repeat this & In Pro Expense, find the 'Club Membership' expense and update it by changing the category to 'Entertainment', setting the amount to 100, and changing the note to 'Monthly fee', then save the changes. & 0.659 \\
\addlinespace[2pt]
14 & Add the expenses from expenses.jpg in Simple Gallery Pro to pro expense. & In Simple Gallery Pro, find the receipt from 'Innovate Solutions Ltd.' and answer what item was purchased and its price. Answer the question in the format: 'Item Name, Price'. & 0.561 \\
\addlinespace[2pt]
15 & Go through the transactions in my\_expenses.txt in Markor. Log the reimbursable transactions in the pro expense. & In the Pro Expense app, find the existing expense record for 'ProDev' and permanently delete it from the logs. & 0.535 \\
\addlinespace[2pt]
16 & Add the following expenses into the pro expense: name\textbar{}amount\_dollars\textbar{}category\_name\textbar{}note Club Membership\textbar{}\$56.67\textbar{}Social\textbar{}Urgent & In Pro Expense, find the 'Club Membership' expense and update it by changing the category to 'Entertainment', setting the amount to 100, and changing the note to 'Monthly fee', then save the changes. & 0.706 \\
\addlinespace[2pt]
17 & Delete all but one of any expenses in pro expense that are exact duplicates, ensuring at least one instance of each unique expense remains. & In the Pro Expense app, find the existing expense record for 'ProDev' and permanently delete it from the logs. & 0.544 \\
\addlinespace[2pt]
18 & Delete all but one of any expenses in pro expense that are exact duplicates, ensuring at least one instance of each unique expense remains. & In the Pro Expense app, find the existing expense record for 'ProDev' and permanently delete it from the logs. & 0.544 \\
\addlinespace[2pt]
19 & Delete the following expenses from pro expense: Textbooks, Salary, Stationery. & In Pro Expense, permanently delete the 'School Supplies' transaction from the recent expenses list. & 0.655 \\
\addlinespace[2pt]
20 & Delete the following expenses from pro expense: Night Out, Tailoring Services, Snacks. & In Pro Expense, permanently delete the 'School Supplies' transaction from the recent expenses list. & 0.617 \\
\addlinespace[2pt]
21 & Delete the following expenses from pro expense: Taxi Fare. & Delete the 'Taxi Fare' transaction from the recent list in the Pro Expense app. & 0.760 \\
\addlinespace[2pt]
22 & Delete the file smart\_guitar\_2023\_07\_08.mp3 from the Android filesystem located in the Podcasts folder within the sdk\_gphone\_x86\_64 storage area. & In Retro Music, permanently delete the audio file for the song 'My Heart is Yours' from the device. & 0.520 \\
\addlinespace[2pt]
23 & Move the file sci\_fi\_thriller.mp4 from Podcasts within the sdk\_gphone\_x86\_64 storage area to the Movies within the same sdk\_gphone\_x86\_64 storage area in the Android filesystem. & In the Files app, navigate to the internal storage 'sdk\_gphone64\_arm64', create a new folder named 'Project\_Alpha', and then change the directory's view mode to the grid layout. & 0.503 \\
\addlinespace[2pt]
24 & Update the Markor note backup\_silly\_tiger.txt by adding the following text, along with a new blank line before the existing content: "rMneE03kV42gdja3JKbM", and rename it to 0xeI\_quick\_house.md. & In the Markor app, open the 'silly\_tree\_copy' note, insert the text 'IMPORTANT:' in bold at the very beginning of the document, and save the file. & 0.668 \\
\addlinespace[2pt]
25 & Update the content of lively\_wolf\_2023\_06\_12.md to "n7g5D1zYaynRyGnucoSD" in Markor and change its name to ucbD\_pretty\_elephant.txt. & In Markor, update the file '2023\_04\_18\_pr...' by replacing its content with a shopping checklist containing bolded 'Apples', italicized 'Bananas', and 'Oranges', then save the changes. & 0.629 \\
\addlinespace[2pt]
26 & Create a new folder in Markor named folder\_20260304\_164543. & Create a new folder named 'Meeting\_Notes' in the Markor app, and inside that folder, create a new file named '2023-10-27\_Briefing.md'. & 0.740 \\
\addlinespace[2pt]
27 & Create a new note in Markor named proud\_banana\_E4ao.txt with the following text: The pen is mightier than the sword. & In the Markor app, open the 'silly\_tree\_copy' note, insert the text 'IMPORTANT:' in bold at the very beginning of the document, and save the file. & 0.605 \\
\addlinespace[2pt]
28 & Create a new note in Markor named wise\_nurse\_2023\_06\_26.txt with the following text: The early bird catches the worm.. Share the entire content of the note with the phone number +11843888619 via SMS using Simple SMS Messenger & In Simple SMS Messenger, save the phone number +16700234643 as a new contact named 'Daily Wisdom' and then initiate a call to this number. & 0.547 \\
\addlinespace[2pt]
29 & Create a note in Markor named final\_fancy\_unicorn.md. Perform a paste operation in the note and save the note. & In Markor, create a new note containing the text 'Draft version', and then use the Search and Replace tool to replace the word 'Draft' with 'Final'. & 0.599 \\
\addlinespace[2pt]
30 & Delete all my notes in Markor. & In the Markor app, delete all files in the Documents folder that contain the phrase 'april\_workout\_routine' in their names. & 0.623 \\
\addlinespace[2pt]
31 & Delete the newest note in Markor. & In the QuickNote section of the Markor app, delete the existing line containing the date '2023-10-15' using the delete line toolbar option. & 0.645 \\
\addlinespace[2pt]
32 & Delete the note in Markor named fancy\_queen\_2023\_01\_05. & In the QuickNote section of the Markor app, delete the existing line containing the date '2023-10-15' using the delete line toolbar option. & 0.700 \\
\addlinespace[2pt]
33 & Edit note\_5v02Y.md in Markor. Add to the bottom of the note The library book is due back on the 15th. & In the QuickNote section of the Markor app, delete the existing line containing the date '2023-10-15' using the delete line toolbar option. & 0.551 \\
\addlinespace[2pt]
34 & Merge the contents of Markor notes best\_lion\_final.txt, alert\_koala\_2023\_10\_10.txt and edited\_super\_cat.md (in the same order) into a new Markor note named QpRyrbS1 and save it. Add a new line between the content of each note. & In Markor, open the file 'tough\_frog\_2023\_08\_05.txt', add a new line with the text 'Urgent' formatted in bold, and then save the file. & 0.587 \\
\addlinespace[2pt]
35 & In Markor, move the note copy\_helpful\_umbrella.txt from MeetingMinutes to WorkProjects. & In the Markor app, add a new entry titled 'Organize documents' to the To-Do list, and then move the file 'final\_meeting\_notes\_project\_team.md' into a new folder named 'Work Archive'. & 0.617 \\
\addlinespace[2pt]
36 & Create a file in Markor, called receipt.md with the transactions from the receipt.png. Use Simple Gallery to view the receipt. Please enter transactions in csv format including the header "Date, Item, Amount". & Create a new file named 'Groceries.md' in Markor, add a checklist item labeled 'Milk', and save the document. & 0.581 \\
\addlinespace[2pt]
37 & Transcribe the contents of video copy\_moment\_10\_.mp4 by watching it in VLC player (located in Download) and writing the sequence of strings shown on each frame to the text file copy\_moment\_10\_\_transcription.txt in Markor as a comma separated list. For example, if the first frame shows the text "edna" and the second frame shows the text "pineapple", then the text file should contain only the following text: "edna, pineapple". & In the Markor app, what is the exact string of characters written on the first line of the file 'oGsN\_note\_X3...'? Answer the question with the text only. & 0.467 \\
\addlinespace[2pt]
38 & Is the note titled 'Research Notes' in the Joplin app marked as a todo item? Respond with either 'True' if it is a todo or 'False' if not. & In the Joplin app, search for the item "Mortgage Payment Schedule" and determine if it is a to-do task (with a checkbox) or a standard note. Answer with "To-do" or "Note". & 0.607 \\
\addlinespace[2pt]
39 & How many attendees were present in the meeting titled 'Marketing Campaign Planning' in the Joplin app? Express your answer as just a single number. & In the Joplin app, how many visible notes contain the word 'Plan' or 'Planning' in their title? Answer the question with a single number. & 0.628 \\
\addlinespace[2pt]
40 & What quantity of matcha powder do I need for the recipe 'Lasagna' in the Joplin app? Express your answer in the format <amount> <unit> where both the amount and unit exactly match the format in the recipe. & In the Broccoli app, for the recipe containing the ingredient 'per individual taste', what is the default serving size displayed when you open the 'Adjust ingredients' dialog? Answer the question with a single number. & 0.502 \\
\addlinespace[2pt]
41 & How many to-dos do I have in the 'Travel' folder in the Joplin app? Express your answer as just a single number. & In the Joplin app's 'All notes' list, how many to-do items start with the word 'Travel'? Answer the question with a single number. & 0.845 \\
\addlinespace[2pt]
42 & Open the camera app. Clear any pop-ups that may appear by granting all permissions that are required. & Grant Android Auto the 'Device \& app notifications' permission to allow it to read notifications, and then clear the app's cache. & 0.528 \\
\addlinespace[2pt]
43 & Add a favorite location marker for Malbun, Liechtenstein in the OsmAnd maps app. & In the OsmAnd app, configure the map settings to display both 'Favorites' locations and 'Transport' routes. & 0.532 \\
\addlinespace[2pt]
44 & Add a location marker for 47.1303814, 9.5930117 in the OsmAnd maps app. & In the OsmAnd app, use the address search feature to locate the point at latitude 40.7306 and longitude -73.9352 using the coordinate search option, and show this location on the map. & 0.579 \\
\addlinespace[2pt]
45 & Save a track with waypoints Schaan, Liechtenstein, Malbun, Liechtenstein, Planken, Liechtenstein, Rotenboden, Liechtenstein in the OsmAnd maps app in the same order as listed. & Using the 'Plan a route' feature in OsmAnd, manually create a custom route by placing four waypoints on the map and save this track to your 'My Places' collection with the name 'Sample Trip'. & 0.606 \\
\addlinespace[2pt]
46 & Add the following recipes into the Broccoli app: title\textbar{}description\textbar{}servings\textbar{}preparationTime\textbar{}ingredients\newline\textbar{}directions Classic Margherita Pizza\textbar{}An ideal recipe for experimenting with different flavors and ingredients.\textbar{}1 serving\textbar{}20 mins\textbar{}to your liking\textbar{}Spread pizza dough with tomato sauce, top with slices of mozzarella cheese and fresh basil leaves. Bake until crust is golden. Garnish with fresh herbs for a more vibrant taste. Garlic Butter Shrimp\textbar{}A quick and easy meal, perfect for busy weekdays.\textbar{}1 serving\textbar{}2 hrs\textbar{}see directions\textbar{}Sauté shrimp in butter and minced garlic until pink. Sprinkle with parsley and serve with lemon wedges. Garnish with fresh herbs for a more vibrant taste. Mango Chicken Curry\textbar{}A delicious and healthy choice for any time of the day.\textbar{}3-4 servings\textbar{}1 hrs\textbar{}various amounts\textbar{}Cook chicken pieces in a pan, add onions, garlic, and ginger. Stir in curry powder, coconut milk, and mango pieces. Simmer until chicken is cooked. Feel free to substitute with ingredients you have on hand. & In the Broccoli app, create a new recipe for 'Classic Margherita Pizza' under the 'Dinner' category, set the description to 'Simple and delicious', source it from 'Chef Mario', specify it serves 2 people, takes 45 minutes to prepare, and list 'Dough, Tomato Sauce, Mozzarella, Basil' as the ingredients. & 0.760 \\
\addlinespace[2pt]
47 & Add the recipes from recipes.jpg in Simple Gallery Pro to the Broccoli recipe app. & In the Broccoli app, create a new recipe for 'Garden Salad', assign it to the 'Healthy' category, and use the device's camera to take and set a cover photo for the recipe before saving. & 0.685 \\
\addlinespace[2pt]
48 & Add the recipes from recipes.txt in Markor to the Broccoli recipe app. & In the Broccoli app, find the 'Tomato Basil Bruschetta' recipe and add it to your Favorites. & 0.595 \\
\addlinespace[2pt]
49 & Add the recipes from recipes.txt in Markor that take 4 hrs to prepare into the Broccoli recipe app. & In the Broccoli app, edit the 'Lentil Soup' recipe to change its preparation time to 45 minutes, and then mark the recipe as a favorite. & 0.662 \\
\addlinespace[2pt]
50 & Add the following recipes into the Broccoli app: Recipe: Caprese Salad Skewers description: A quick and easy meal, perfect for busy weekdays. servings: 6 servings preparationTime: 4 hrs ingredients: various amounts directions: Thread cherry tomatoes, basil leaves, and mozzarella balls onto skewers. Drizzle with balsamic glaze. Garnish with fresh herbs for a more vibrant taste. & In the Broccoli app, the 'Caprese Salad Skewers' recipe is missing ingredient details. Edit the recipe to add 'Mozzarella balls' and 'Cherry tomatoes' to the ingredients list, and update the preparation time to '20 mins'. & 0.780 \\
\addlinespace[2pt]
51 & Delete all but one of any recipes in the Broccoli app that are exact duplicates, ensuring at least one instance of each unique recipe remains & In the Broccoli app, clean up the recipe list by deleting the duplicate entries for 'Baked Cod with Lemon and Dill' so that only one such entry remains. & 0.771 \\
\addlinespace[2pt]
52 & Delete all but one of any recipes in the Broccoli app that are exact duplicates, ensuring at least one instance of each unique recipe remains & In the Broccoli app, clean up the recipe list by deleting the duplicate entries for 'Baked Cod with Lemon and Dill' so that only one such entry remains. & 0.771 \\
\addlinespace[2pt]
53 & Delete all but one of any recipes in the Broccoli app that are exact duplicates, ensuring at least one instance of each unique recipe remains & In the Broccoli app, clean up the recipe list by deleting the duplicate entries for 'Baked Cod with Lemon and Dill' so that only one such entry remains. & 0.771 \\
\addlinespace[2pt]
54 & Delete the following recipes from Broccoli app: Eggplant Parmesan, Cauliflower Fried "Rice", Lemon Garlic Tilapia. & Delete the 'Eggplant Parmesan' recipe from your collection in the Broccoli app. & 0.771 \\
\addlinespace[2pt]
55 & Delete the recipes from Broccoli app that use ghee in the directions. & Delete the 'Beef Stir Fry' recipe from the Broccoli app. & 0.682 \\
\addlinespace[2pt]
56 & Delete the following recipes from Broccoli app: Turkey and Cheese Panini, Stuffed Bell Peppers, Thai Peanut Noodle Salad. & Delete the 'Thai Peanut Noodle Salad' recipe from the Broccoli app. & 0.796 \\
\addlinespace[2pt]
57 & Delete the following recipes from Broccoli app: Chicken Caesar Salad Wrap. & Delete the 'Beef Stir Fry' recipe from the Broccoli app. & 0.720 \\
\addlinespace[2pt]
58 & Delete the following recipes from Broccoli app: Mango Chicken Curry. & Delete the 'Beef Stir Fry' recipe from the Broccoli app. & 0.752 \\
\addlinespace[2pt]
59 & Create a playlist in Retro Music titled "Acoustic Sessions 86" with the following songs, in order: City of Stars, Echoes of Silence & In Retro Music, create a new playlist named 'Acoustic Sessions' and add the songs 'City of Stars' and 'Distant Memories' to it. & 0.829 \\
\addlinespace[2pt]
60 & Add the following songs, in order, Shadows of Time, Eternal Flame, Golden Days to my playing queue in Retro music. & In Retro Music, find the song 'Distant Memories' in the 'Last added' list and add it to the playing queue. & 0.648 \\
\addlinespace[2pt]
61 & Create a playlist in Retro Music titled "Electronic Chillout 553" with a duration between 45 and 50 minutes using the provided songs. & In Retro Music, create a new playlist named 'Chill Vibes' and add the song 'Hidden Paths' to it. & 0.681 \\
\addlinespace[2pt]
62 & Create a playlist in Retro Music titled "Retro Pop Hits 458" with the following songs, in order: Bright Lights, Eternal Flame, Endless Summer. Then export the playlist to the Downloads directory on the device. & Create a new playlist named 'Night Drive' in the Retro Music app and add the songs 'Bright Lights' by Oliver and 'Eternal Flame' by Martina to it. & 0.753 \\
\addlinespace[2pt]
63 & In Simple Gallery Pro, copy receipt\_smart\_vase\_copy.jpg in DCIM and save a copy with the same name in Download & In Simple Gallery Pro, locate the receipt image for 'Innovate Solutions Ltd' within the DCIM folder, rotate the image 90 degrees clockwise, and mark it as a favorite. & 0.647 \\
\addlinespace[2pt]
64 & In Simple Calendar Pro, create a calendar event on 2023-10-17 at 11h with the title 'Review session for Campaign' and the description 'We will review product launch. Snacks will be provided.'. The event should last for 45 mins. & In Simple Calendar Pro, create a new event on October 25 titled 'Budget Review' at 'Finance Dept' that starts at 10:00 and ends at 11:30, and add the note 'Prepare Q3 reports' in the description. & 0.770 \\
\addlinespace[2pt]
65 & In Simple Calendar Pro, create a calendar event in two weeks from today at 20h with the title 'Workshop on Annual Report' and the description 'We will discuss upcoming project milestones.'. The event should last for 60 mins. & In Simple Calendar Pro, create a new event titled "Strategy Workshop" for October 20th starting at 14:00, set the location to "Main Hall", and add a description "Quarterly planning session". & 0.767 \\
\addlinespace[2pt]
66 & In Simple Calendar Pro, create a calendar event for this Thursday at 5h with the title 'Call with HR' and the description 'We will discuss annual budget. Looking forward to productive discussions.'. The event should last for 45 mins. & In Simple Calendar Pro, create a new event on October 25 titled 'Budget Review' at 'Finance Dept' that starts at 10:00 and ends at 11:30, and add the note 'Prepare Q3 reports' in the description. & 0.748 \\
\addlinespace[2pt]
67 & In Simple Calendar Pro, create a calendar event for tomorrow at 0h with the title 'Appointment for Campaign' and the description 'We will celebrate software updates.'. The event should last for 45 mins. & In Simple Calendar Pro, create a new event titled 'Project Meeting' for tomorrow at 2:00 PM, set it to repeat weekly, and add a reminder 10 minutes before the start. & 0.748 \\
\addlinespace[2pt]
68 & In Simple Calendar Pro, create a recurring calendar event titled 'Review session for Project X' starting on 2023-10-24 at 18h. The event recurs weekly, forever, and lasts for 45 minutes each occurrence. The event description should be 'We will organize team roles. Let's be punctual.'. & In Simple Calendar Pro, create a new event titled 'Review session for Annual Report' for October 21st. Set the description to 'We will organize annual budget. Let's be punctual.' and configure the event to repeat yearly. & 0.788 \\
\addlinespace[2pt]
69 & Do I have any events October 16 2023 in Simple Calendar Pro? Answer with the titles only. If there are multiples titles, format your answer in a comma separated list. & In Simple Calendar Pro, identify all events scheduled for October 17th and answer with their titles separated by a comma. & 0.782 \\
\addlinespace[2pt]
70 & In Simple Calendar Pro, delete all the calendar events on 2023-10-25 & In Simple Calendar Pro, find and delete the 'Workshop' event scheduled for October 24th. & 0.718 \\
\addlinespace[2pt]
71 & In Simple Calendar Pro, delete all events scheduled for this Friday. & In Simple Calendar Pro, find and delete the 'Workshop' event scheduled for October 24th. & 0.727 \\
\addlinespace[2pt]
72 & In Simple Calendar Pro, delete the calendar event on 2023-10-30 at 11h with the title 'Catch up on Annual Report' & In Simple Calendar Pro, find and delete the 'Workshop' event scheduled for October 24th. & 0.751 \\
\addlinespace[2pt]
73 & What is on my schedule for October 19 2023 at 21:45 in Simple Calendar Pro? Answer with the titles only. If there are multiples titles, format your answer in a comma separated list. & In Simple Calendar Pro, switch the calendar view to 'Monthly and daily view', and then list the titles of all events scheduled for October 23rd. Answer with the titles separated by a comma. & 0.757 \\
\addlinespace[2pt]
74 & What events do I have in the next week in Simple Calendar Pro? Assume the week starts from Monday. Answer with the titles only. If there are multiples titles, format your answer in a comma separated list. & In Simple Calendar Pro, identify all events scheduled for October 17th and answer with their titles separated by a comma. & 0.750 \\
\addlinespace[2pt]
75 & Do I have any events between 4pm and 8pm October 27 2023 in Simple Calendar Pro? Answer with the titles only. If there are multiples titles, format your answer in a comma separated list. & In Simple Calendar Pro, identify all events scheduled for October 17th and answer with their titles separated by a comma. & 0.744 \\
\addlinespace[2pt]
76 & What events do I have October 17 2023 in Simple Calendar Pro? Answer with the titles only. If there are multiple titles, format your answer as a comma separated list. & In Simple Calendar Pro, identify all events scheduled for October 17th and answer with their titles separated by a comma. & 0.831 \\
\addlinespace[2pt]
77 & What is my first event after 11:00am October 21 2023 in Simple Calendar Pro? Answer with the titles only. If there are multiples titles, format your answer in a comma separated list. & In Simple Calendar Pro, switch the calendar view to 'Monthly and daily view', and then list the titles of all events scheduled for October 23rd. Answer with the titles separated by a comma. & 0.773 \\
\addlinespace[2pt]
78 & What is the location of my Coding challenge event in Simple Calendar Pro? Answer with the location only. & In Simple Calendar Pro, what is the description for the 'Call with the Team' event, and is there a location currently set for it?. & 0.597 \\
\addlinespace[2pt]
79 & What is my next upcoming event in Simple Calendar Pro? Answer with the title only. If there are multiples titles, format your answer in a comma separated list. & In Simple Calendar Pro, identify all events scheduled for October 17th and answer with their titles separated by a comma. & 0.753 \\
\addlinespace[2pt]
80 & When is my next meeting with Muhammad in Simple Calendar Pro? Express your answer in the format <month name> <day> <year> <hour in 24-hour format>:<minutes>. & In Simple Calendar Pro, switch the view to the 'Simple event list', find the 'Meeting with Marketing' event, and answer with its start time in HH:MM format. & 0.629 \\
\addlinespace[2pt]
81 & Create a new drawing in Simple Draw Pro. Name it amet\_lively\_eagle\_final.jpg. Save it in the Pictures folder within the sdk\_gphone\_x86\_64 storage area. & Using Simple Draw Pro, save a file named 'system\_diagram' directly to the root directory of the internal storage (sdk\_gphone64\_arm64). & 0.753 \\
\addlinespace[2pt]
82 & Reply to +13431223053 with message: Actions speak louder than words. in Simple SMS Messenger & In Simple SMS Messenger, find the conversation containing the message 'Actions speak louder than words.', and add the sender to your device's contacts with the first name 'Wisdom'. & 0.669 \\
\addlinespace[2pt]
83 & Reply to the most recent text message using Simple SMS Messenger with message: When in Rome, do as the Romans do. & Find and delete the conversation thread containing the message "When in Rome, do as the Romans do." in the Simple SMS Messenger app. & 0.621 \\
\addlinespace[2pt]
84 & Resend the message I just sent to Lily Pereira in Simple SMS Messenger & In Simple SMS Messenger, resume the draft conversation with +17401638798 by appending ' hope you are well' to the existing text and sending it. & 0.572 \\
\addlinespace[2pt]
85 & Send a text message using Simple SMS Messenger to +15039078312 with message: Lorem Ipsum is simply dummy text. & Using Simple SMS Messenger, send a text message with the content 'padiNoBMVR' to the number +1 545-178-61614309. & 0.631 \\
\addlinespace[2pt]
86 & Send a message to +17228051441 with the clipboard content in Simple SMS Messenger & In Simple SMS Messenger, use the dialer to enter the number +1 545-178-6161 and then start a new text message to this recipient. & 0.571 \\
\addlinespace[2pt]
87 & Text the address of the event to David Wang that Emily Liu just sent me in Simple SMS Messenger & In Simple SMS Messenger, send a text message to +1 930-572-4145+2 with the content 'Please confirm if the meeting is still on for today'. & 0.546 \\
\addlinespace[2pt]
88 & How many skate boarding activities did I do this week in the OpenTracks app? Assume the week starts from Monday. Express your answer as a single integer. & In the OpenTracks app, how many activities shown in the list were recorded on a Thursday? Answer with a single number. & 0.718 \\
\addlinespace[2pt]
89 & What activities did I do October 6 2023 in the OpenTracks app? Answer with the activity type only. If there are multiple types, format your answer in a comma separated list. & In the OpenTracks app, identify all activities that have a recorded distance greater than 9 miles. Answer with the names of the activities separated by a comma. & 0.693 \\
\addlinespace[2pt]
90 & How long was my climbing activity October 15 2023 in the OpenTracks app? Express your answer in minutes as a single integer. & In the OpenTracks app, what are the moving time and elevation gain recorded for the 'Morning Run' activity? Answer with the values separated by a comma. & 0.611 \\
\addlinespace[2pt]
91 & What was the longest distance covered in a kayaking activity in the OpenTracks app this week? Assume the week starts from Monday. Express your answer as a single number in meters rounded to the nearest integer. & In the OpenTracks app list, how many recorded activities have a distance greater than 10 miles? Answer with a single number. & 0.666 \\
\addlinespace[2pt]
92 & What was the total distance covered for swimming activities in the OpenTracks app from October 6 2023 to October 15 2023? Express your answer as a single number in meters rounded to the nearest integer. & In the OpenTracks app list, how many recorded activities have a distance greater than 10 miles? Answer with a single number. & 0.659 \\
\addlinespace[2pt]
93 & What was the total duration of hiking activities in the OpenTracks app this week? Assume the week starts from Monday. Express your answer in minutes as a single integer. & In the OpenTracks app, how many activities shown in the list were recorded on a Thursday? Answer with a single number. & 0.618 \\
\addlinespace[2pt]
94 & Turn bluetooth off. & Turn off the Nearby Share feature completely, then return to the Connection preferences menu and open the Bluetooth settings. & 0.638 \\
\addlinespace[2pt]
95 & Turn bluetooth off. & Turn off the Nearby Share feature completely, then return to the Connection preferences menu and open the Bluetooth settings. & 0.638 \\
\addlinespace[2pt]
96 & Turn bluetooth on. & Navigate to the Bluetooth pairing screen in Settings and find the phone's Bluetooth address. Answer the question with the full address string. & 0.515 \\
\addlinespace[2pt]
97 & Turn bluetooth on. & Navigate to the Bluetooth pairing screen in Settings and find the phone's Bluetooth address. Answer the question with the full address string. & 0.515 \\
\addlinespace[2pt]
98 & Turn brightness to the max value. & Configure the device display for better visibility by enabling 'Dark theme' and setting the 'Font size' to the largest available option. & 0.403 \\
\addlinespace[2pt]
99 & Turn brightness to the max value. & Configure the device display for better visibility by enabling 'Dark theme' and setting the 'Font size' to the largest available option. & 0.403 \\
\addlinespace[2pt]
100 & Turn brightness to the min value. & Configure the device display for better visibility by enabling 'Dark theme' and setting the 'Font size' to the largest available option. & 0.316 \\
\addlinespace[2pt]
101 & Turn brightness to the min value. & Configure the device display for better visibility by enabling 'Dark theme' and setting the 'Font size' to the largest available option. & 0.316 \\
\addlinespace[2pt]
102 & Copy the following text to the clipboard: Reservation under: Mike & In the Markor app, append the sentence ' Dinner reserved at 7pm.' to the existing text in the file named '2023\_08\_11\_good\_vase.txt'. & 0.375 \\
\addlinespace[2pt]
103 & Turn wifi off. & Turn off the "Adaptive connectivity" feature and set the Private DNS mode to "Off". & 0.510 \\
\addlinespace[2pt]
104 & Turn wifi off. & Turn off the "Adaptive connectivity" feature and set the Private DNS mode to "Off". & 0.510 \\
\addlinespace[2pt]
105 & Turn wifi on. & Set up a portable Wi-Fi hotspot named 'TravelRouter' with the password 'SecureNet99', and turn it on to share your cellular internet connection. & 0.488 \\
\addlinespace[2pt]
106 & Turn wifi on. & Set up a portable Wi-Fi hotspot named 'TravelRouter' with the password 'SecureNet99', and turn it on to share your cellular internet connection. & 0.488 \\
\addlinespace[2pt]
107 & Which tasks have I completed for October 18 2023 in Tasks app? Answer with the titles only. If there are multiples titles, format your answer in a comma separated list. & In the Tasks app, identify the titles of all tasks that are specifically due on 'Oct 8'. Answer by listing the titles separated by a comma. & 0.782 \\
\addlinespace[2pt]
108 & How many tasks do I have due next week in Tasks app? Assume the week starts from Monday. Express your answer as a single integer. & In the Tasks app, count the number of visible tasks that are due on 'Tue'. Answer with the single number. & 0.715 \\
\addlinespace[2pt]
109 & What tasks do I have due October 21 2023 in Tasks app? Answer with the titles only. If there are multiples titles, format your answer in a comma separated list. & In the Tasks app, identify the titles of all tasks that are specifically due on 'Oct 8'. Answer by listing the titles separated by a comma. & 0.773 \\
\addlinespace[2pt]
110 & What are my high priority tasks in Tasks app? Answer with the titles only. If there are multiples titles, format your answer in a comma separated list. & In the Tasks app, how many tasks are currently active (not completed), and which of them has the highest priority? Answer the question with the number and task title, separated by a comma. & 0.759 \\
\addlinespace[2pt]
111 & Which tasks with high priority are due October 16 2023 in the Tasks app? Answer with the title only. If there are multiples titles, format your answer in a comma separated list. & In the Tasks app, identify the titles of all tasks that are specifically due on 'Oct 8'. Answer by listing the titles separated by a comma. & 0.779 \\
\addlinespace[2pt]
112 & What incomplete tasks do I have still have to do by October 21 2023 in Tasks app? Answer with the titles only. If there are multiples titles, format your answer in a comma separated list. & In the Tasks app, identify the titles of all tasks that are specifically due on 'Oct 8'. Answer by listing the titles separated by a comma. & 0.740 \\
\addlinespace[2pt]
113 & Turn off WiFi, then enable bluetooth & Turn off the Nearby Share feature completely, then return to the Connection preferences menu and open the Bluetooth settings. & 0.581 \\
\addlinespace[2pt]
114 & Turn on Wifi, then open the contacts app & Open the Contacts app and determine how many contacts are currently saved in the list. Answer with a single number. & 0.516 \\
\addlinespace[2pt]
115 & Create a playlist titled "Documentary Insights Favorites" with the following files in VLC (located in Internal Memory/VLCVideos), in order: 2023\_01\_29\_episode\_46\_HD.mp4, 2023\_06\_29\_clip\_38\_export.mp4 & In the VLC app, locate the 'Documents' folder within the internal memory storage and add it to the 'Favorites' list. & 0.615 \\
\addlinespace[2pt]
116 & Create a playlist titled "Recipe Collection Ultimate Collection" with the following files in VLC (located in Internal Memory/VLCVideos), in order: moment\_95\_\_1KUB.mp4, scene\_54\_raw\_gbYs.mp4, moment\_52\_HD\_final.mp4, highlight\_13\_HD\_2023\_01\_29.mp4. And then, create a playlist titled "Ultimate Fails Ultimate Collection" with the following files in VLC, in order: recording\_41\_HD\_backup.mp4, recording\_56\_\_JRVN.mp4, 2023\_08\_10\_scene\_27\_raw.mp4. & In the VLC app, navigate to the Browse tab and create a new playlist named "My Top Hits" using all the media files found in the "Music" folder. & 0.619 \\
\end{longtable}
\end{scriptsize}

\end{document}